\documentclass[fleqn,10pt]{wlscirep}
\usepackage[utf8]{inputenc}
\usepackage[T1]{fontenc}
\usepackage{subfigure}
\usepackage[percent]{overpic}
\newtheorem{thm}{Proposition}
\usepackage{float}
\usepackage{hyperref}

\title{ODBAE: a high-performance model identifying complex phenotypes in high-dimensional biological datasets}

\author[1,2,+]{Yafei Shen}
\author[3,+]{Tao Zhang}
\author[4]{Zhiwei Liu}
\author[6]{Kalliopi Kostelidou}
\author[4,5,*]{Ying Xu}
\author[1,2,*]{Ling Yang}

\affil[1]{School of Mathematical Sciences, Soochow University, Suzhou, 215006, Jiangsu, China}
\affil[2]{Center for Systems Biology, Soochow University, Suzhou, 215006, Jiangsu, China}
\affil[3]{Jiangsu Province Engineering Research Center of Development and Translation of Key Technologies for Chronic Disease Prevention and Control, Suzhou Vocational Health College, Suzhou, 215009, Jiangsu, China}
\affil[4]{Cambridge-Suda Genomic Research Center, Suzhou medical college of Soochow University, Suzhou, 215123, Jiangsu, China}
\affil[5]{Jiangsu Key Laboratory of Neuropsychiatric Diseases, Suzhou medical college of Soochow University, Suzhou, 215123, Jiangsu, China}
\affil[6]{KTL Research Acceleration LTD, Lapithou 11, 2410, Nicosia, Cyprus}
\affil[+]{These authors contributed equally}
\affil[*]{e-mail: lyang@suda.edu.cn (Ling Yang); yingxu@suda.edu.cn (Ying Xu)}

\begin{abstract}
Identifying complex phenotypes from high-dimensional biological data is challenging due to the intricate interdependencies among different physiological indicators. Traditional approaches often focus on detecting outliers in single variables, overlooking the broader network of interactions that contribute to phenotype emergence. Here, we introduce ODBAE (Outlier Detection using Balanced Autoencoders), a machine learning method designed to uncover both subtle and extreme outliers by capturing latent relationships among multiple physiological parameters. ODBAE's revised loss function enhances its ability to detect two key types of outliers: influential points (IP), which disrupt latent correlations between dimensions, and high leverage points (HLP), which deviate from the norm but go undetected by traditional autoencoder-based methods. Using data from the International Mouse Phenotyping Consortium (IMPC), we show that ODBAE can identify knockout mice with complex, multi-indicator phenotypes - normal in individual traits, but abnormal when considered together. In addition, this method reveals novel metabolism-related genes and uncovers coordinated abnormalities across metabolic indicators. Our results highlight the utility of ODBAE in detecting joint abnormalities and advancing our understanding of homeostatic perturbations in biological systems.
\end{abstract}
\begin{document}

\flushbottom
\maketitle
%
%
\thispagestyle{empty}


\section*{Introduction}
Phenotypes, including symptoms, defined as the observable characteristics or traits of an organism, are the result of complex interactions between genotype and environmental factors\cite{Kurbatova2015PhenStatAT,Ar2015FindingOW}. These traits are often quantified using physiological or pathological indicators that serve as proxies for the underlying biological or pathological processes. Recent advances in phenotypic research, particularly through genetic manipulation have deepen our understanding of gene function and the mechanisms underlying pathological conditions across a wide range of organisms, including plants and animals\cite{Ar2015FindingOW}. Traditionally, gene-related phenotypic analyses have focused on identifying abnormalities in individual physiological indicators\cite{Rozman2018IdentificationOG,Swan2020MouseMP,Meehan2017DiseaseMD}. However, findings from the International Mouse Phenotyping Consortium (IMPC) demonstrate strong correlations between physiological indicators, suggesting that many traits and diseases result not from isolated abnormalities, but from coordinated disruptions across multiple physiological indicators\cite{Nicholson2022MultivariatePA}.

Our current understanding of phenotypes or diseases often emphasizes the detection of outliers in individual physiological indicators, typically those outside the normal physiological range. However, the emergence of disease in an organism is a more complex process. Homeostasis-the dynamic balance maintained by biological systems-can be perturbed at multiple levels before a single indicator deviates outside the normal range\cite{Banfalvi2013HomeostasisT}. This suggests that phenotypic abnormalities may manifest as imbalances between correlated indicators, even when each individual measure remains within its expected range. These imbalances may represent early warning signs of disease or dysfunction that may be missed by traditional univariate analysis.

We hypothesize that the coordinated perturbation of physiological indicators reflects an emerging phenotype or disease state, even when individual measures appear normal. By examining these subtle interdependencies, we can gain insight into how homeostasis is perturbed in knockout mouse models. Furthermore, for genes whose knockout does not result in obvious outliers in individual parameters, disruption of homeostatic balance may still occur through correlated indicators. This approach challenges the notion that the absence of a detectable abnormal phenotype in knockout mice indicates no functional impact, and instead suggests that our focus on single-indicator abnormalities may obscure more complex systemic effects.

Existing methods, such as linear regression\cite{Nicholson2022MultivariatePA,Elgart2022CorrelationsBC} and basic bioinformatics tools\cite{BulikSullivan2015AnAO,Kanai2018GeneticAO,Lyu2021GESLMAF}, have provided initial insights into these relationships. However, these techniques struggle to capture non-linear and complex relationships between multiple physiological indicators, especially in high-dimensional biological datasets. Machine learning methods, particularly autoencoders, have emerged as powerful tools for detecting outliers by learning complex, latent patterns within data\cite{Sattarov2022ExplainingAU,Lu2017UnsupervisedSO,Wei2022LSTMAutoencoderBasedAD,Ding2019DeepAD,Chen2019UnsupervisedCO,Xu2015LearningDR,Le2022AttentionbasedRA}. By compressing and reconstructing data, autoencoders can capture subtle variations and relationships that may indicate phenotypic abnormalities across multiple indicators. Their ability to learn from high-dimensional data makes them well suited for analyzing the intricate dynamics involved in phenotypic expression\cite{Zhou2019OptimizingAF,Liu2022IntegrationOH}. In the regression analysis, all outliers can be categorized as influential points (IP) and high leverage points (HLP). IP exerts a substantial influence on the model’s fitting results and HLP deviate from the center of the dataset but may not necessarily impact the model fit\cite{Imon2013IdentificationOM}. Since the reconstruction of any dataset by the autoencoder can be regarded as a process of regression, IP can be easily detected when using autoencoders for outlier detection. However, there are still many uncertainties about the detection effect of HLP.

In this study, we introduce ODBAE (Outlier Detection using Balanced Autoencoders), a machine learning method designed to detect outliers by capturing the relationships between physiological indicators. ODBAE improves on traditional autoencoder by balancing the reconstruction error across principal component directions, enhancing its ability to detect both IP that deviate from the expected relationships between indicators and HLP that are far from the data center. We apply ODBAE to developmental and metabolic datasets from IMPC to explore how coordinated perturbations in physiological parameters can reveal previously unidentified phenotypes, even in cases where knockout mice show no abnormalities in individual traits. By using ODBAE, we aim to identify novel gene functions and phenotypic complexities that have been missed by single-indicator screening methods.

\section*{Results}
\subsection*{Outlier detection in ODBAE}
ODBAE takes tabular datasets from various sources as input, such as gene knockout mouse dataset from IMPC and patient health examination data. Each row in the data represents a record, and each column represents an attribute. The goal of ODBAE is to maintain the effectiveness of the autoencoder in detecting IP while further improving the detection performance for HLP. This is a comprehensive approach to outlier detection in high-dimensional tabular datasets. Furthermore, ODBAE also provides an anomaly explanation for each outlier, indicating which parameter or parameters are associated with the anomaly. For gene knockout mouse datasets, ODBAE can identify genes that, when knocked out, lead to anomalies, as well as the abnormal parameter pairs associated with the gene knockout.

\subsection*{Overview of ODBAE}
\begin{figure}[htbp]%
	\centering
	\subfigure{
		\label{overview_of_ODBAE}
		\begin{overpic}[width=0.8\textwidth]{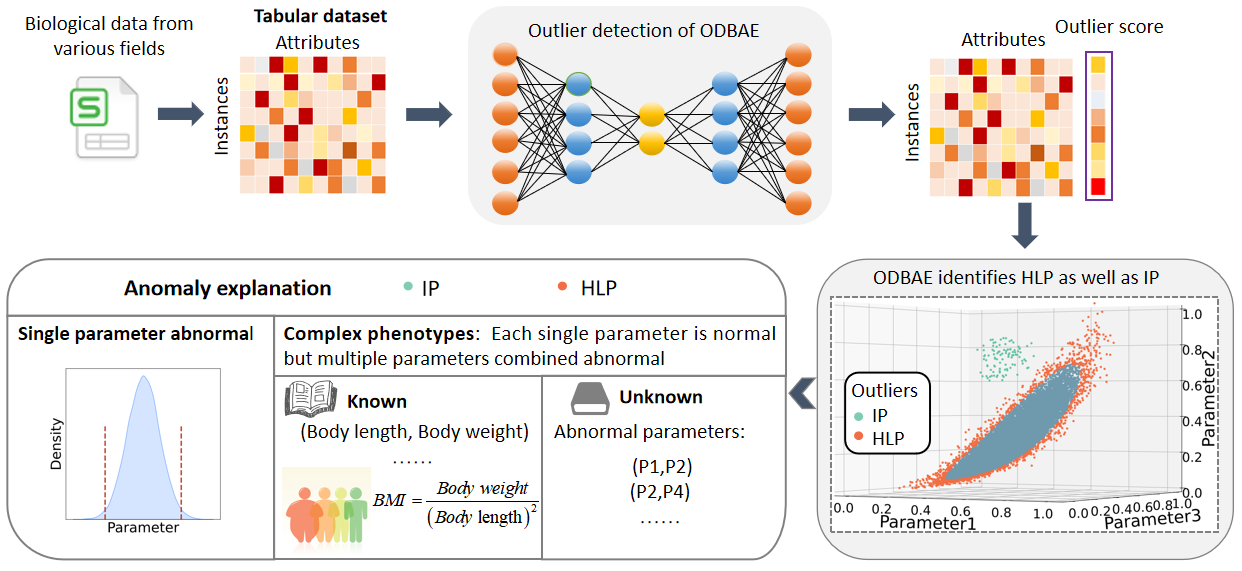}
			\put(-3,46){\large\textbf{a}}
	\end{overpic}}
	
	\vspace{0.2in}
	\subfigure{
		\label{overview_plot4}
		\begin{overpic}[width=0.8\textwidth]{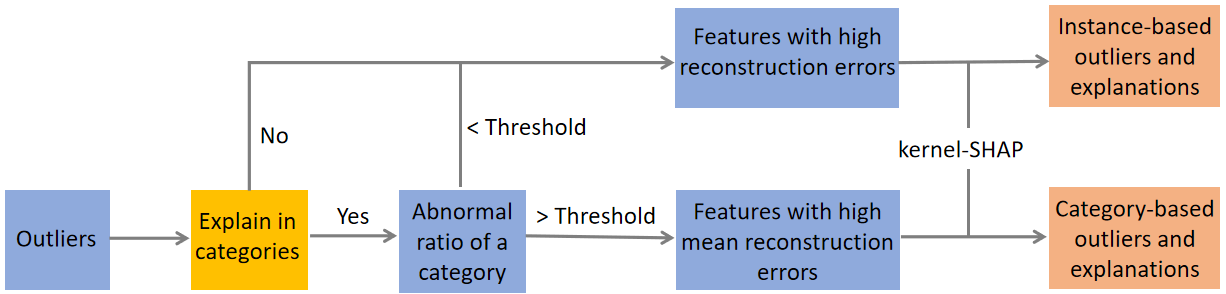}
			\put(-3,24){\large\textbf{b}}
		\end{overpic}
	}
	
	\vspace{0.2in}
	\subfigure{
		\label{append_fig4}
		\begin{overpic}[width=0.8\textwidth]{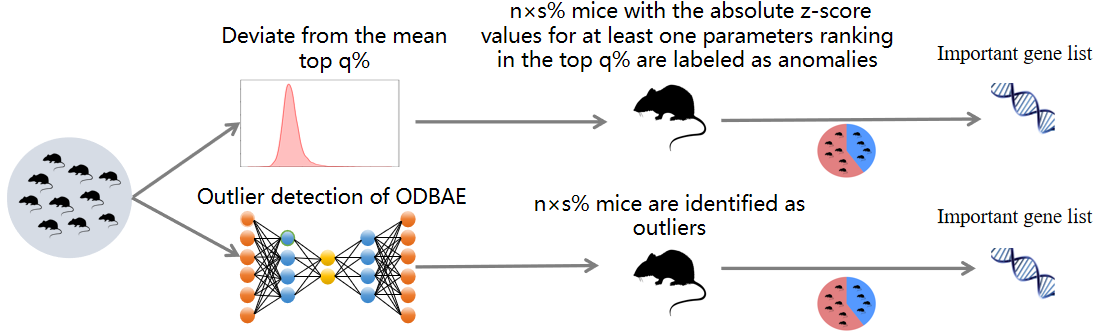}
			\put(-3,25){\large\textbf{c}}
		\end{overpic}
	}
	
	\vspace{0.2in}
	\subfigure{
		\label{overview_plot6}
		\begin{overpic}[width=0.8\textwidth]{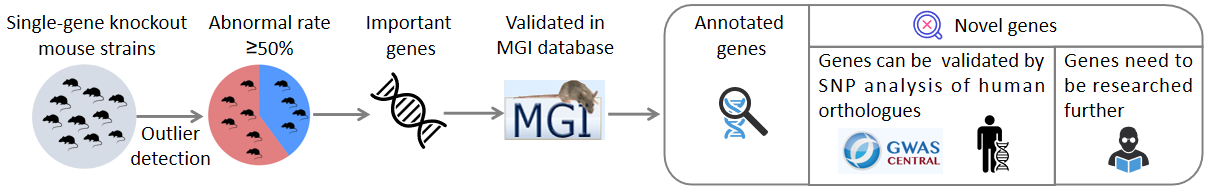}
			\put(-3,15){\large\textbf{d}}
	\end{overpic}}
	\caption{{\bf ODBAE is an approach for identifying and explaining outliers in tabular datasets, it can discover complex phenotypes and identify novel genes when applied to knockout mouse datasets.} {\bf a}, Process for discovering complex phenotypes. ODBAE first detect outliers based on a balanced autoencoder which can balance the reconstruction of the training dataset to detect both HLP and IP, then perform anomaly explanation to gain abnormal parameters. Finally, some outliers are identified due to joint anomalies of multiple parameters. {\bf b}, Anomaly explanation of ODBAE. The detected outliers are explained in terms of SHAP values. {\bf c}, For both males and females, abnormal mice are first identified according to the presence of metabolic parameters that deviated too far from the mean, and then the same abnormal proportion is set to identify abnormal mice with ODBAE. Finally, some important genes will only be obtained by ODBAE. {\bf d}, Genes with an abnormal proportion of more than $50$\% of the corresponding single-gene knockout mouse strains are considered important genes by ODBAE. If an important gene has no known metabolic links in the MGI database, it will be considered a novel gene and further validated by SNP analysis of human orthologues. }\label{fig1}
\end{figure}

To address the challenge of outlier detection in high-dimensional tabular datasets, we construct an improved autoencoder model (ODBAE) with refined training loss function. Our model leverages the strong representation learning and feature extraction capabilities of traditional autoencoders for IP, while mitigating their limitations in HLP detection (Methods). The core principle underlying ODBAE-based outlier detection is that inliers can be well reconstructed, whereas outliers, including both HLP and IP, may generate significant reconstruction errors. ODBAE performs unsupervised outlier detection on the input high-dimensional datasets. During training process, ODBAE learns as much intrinsic information as possible from training set and reconstructs the training dataset by minimizing the loss function. The training dataset selection strategy depends on the initial outlier prevalence within the dataset. For datasets with few outliers presence, the entire dataset are utilized for both training and testing for outlier detection, such as data from wild-type mice in IMPC. However, if the overall proportion of outliers are not clear, a subset of the data exhibiting fewer anomalies is chosen for model training. Then, the entire dataset, or other subset, can be used for outlier detection. Subsequently, the trained model is applied to the test dataset, generating reconstruction error of all sample points. Finally, sample points with reconstruction errors greater than the predefined threshold are considered outliers. As shown in Fig.\ref{fig1}a, outlier detection from biological datasets by ODBAE include three steps: (1) given a training dataset as input, the intrinsic information of normal data points are learned; (2) the trained model is used to reconstruct the test dataset, and outliers will be identified according to large reconstruction error; (3) the detected outliers including HLP and IP are explained using highest reconstruction errors and kernel-SHAP to gain the abnormal parameters\cite{Antwarg2021ExplainingAD}.

Based on mathematical analysis, ODBAE uses a revised training loss function incorporating an appropriate penalty term to Mean Square Error (MSE) to balance the reconstruction by properly suppressing complete reconstruction of the autoencoder (Methods). Specifically, the penalty term can ensure the equal eigenvalue difference between each principal component direction of the training and reconstructed dataset. Therefore, the incorporated penalty term enhance the detection of HLP, while the MSE term maintain the identification performance of IP.

To explain each outlier detected by ODBAE, we first identify the top features that contribute the most to the reconstruction error, and then apply kernel-SHAP to obtain the features that have the greatest impact on them (Fig.\ref{fig1}b and Methods). Finally, ODBAE outputs instance-based outliers and their explanations. For outliers in categories, if the anomaly rate for a category of outliers is greater than the set threshold, then ODBAE provide anomaly explanation for each category according to their mean values of each feature.

\subsection*{ODBAE's comprehensive benefits}
The power of ODBAE lies in its ability to efficiently detect both HLP and IP in complex, high-dimensional biological data sets. HLP are outliers that deviate from the central distribution, while IP significantly influence model results. To evaluate the performance of ODBAE in detecting IP, we compared it to principal component analysis (PCA)\cite{2003A}, a common outlier detection method. ODBAE demonstrated superior performance, particularly in identifying IP, due to its enhanced ability to capture nonlinear relationships between data points (Supplementary Fig.1).

We then applied ODBAE to phenotypic data from the IMPC to identify complex phenotypes in knockout mouse models. Using a threshold of the top 2\% of absolute z-scores\cite{Shalabi2006DataMA} for any given physiological parameter, we flagged mice as potential outliers (Methods). We then applied ODBAE with a predefined abnormality ratio (Fig.\ref{fig1}c). If more than 50\% of mice from a single-gene knockout strain were classified as outliers, the corresponding gene was identified as significant and further analyzed (Fig.\ref{fig1}d). ODBAE was able to identify key mutant strains with complex phenotypes, even in cases where individual physiological indicators appeared normal.

\begin{figure}[htbp]
	\centering
	\subfigure{
		\label{section3_BMI_example}  
		\begin{overpic}[width=0.28\textwidth]{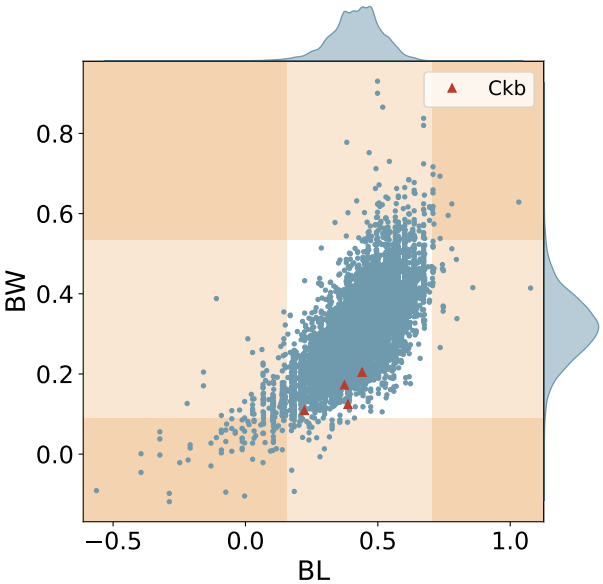}
			\put(0,82){\large\textbf{a}}
	\end{overpic}}
	\hspace{0.2in} %
	\subfigure{
		\label{section3_BMI_example_BMI_distribution}
		\begin{overpic}[width=0.42\textwidth]{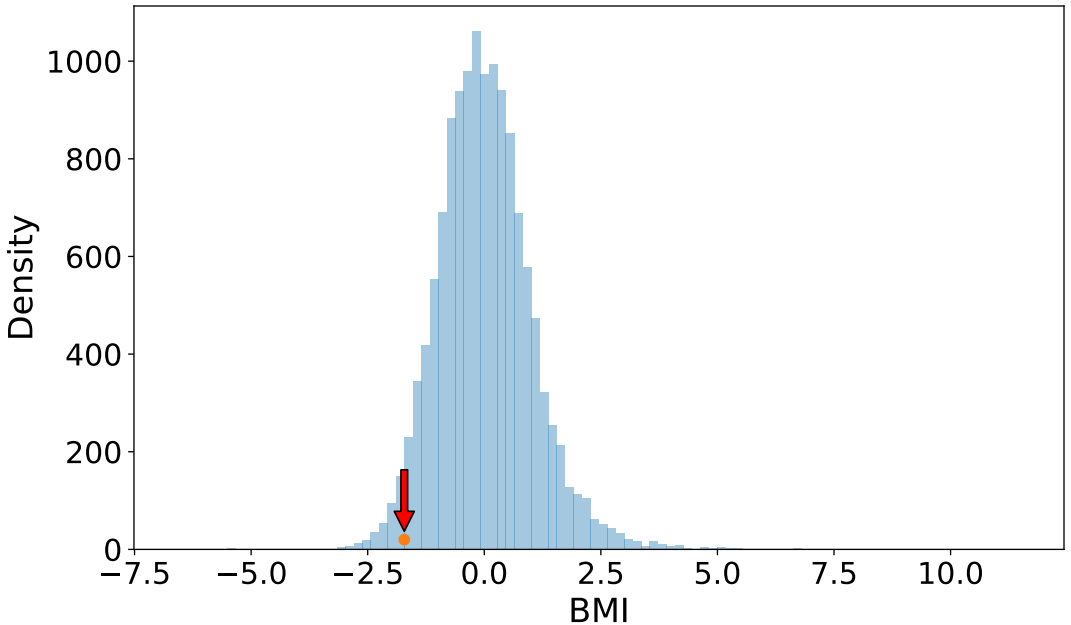}
			\put(2,55){\large\textbf{b}}
	\end{overpic}}
	
	\subfigure{
		\label{section1_generate_example1_mse_loss1000}
		\begin{overpic}[width=0.24\textwidth]{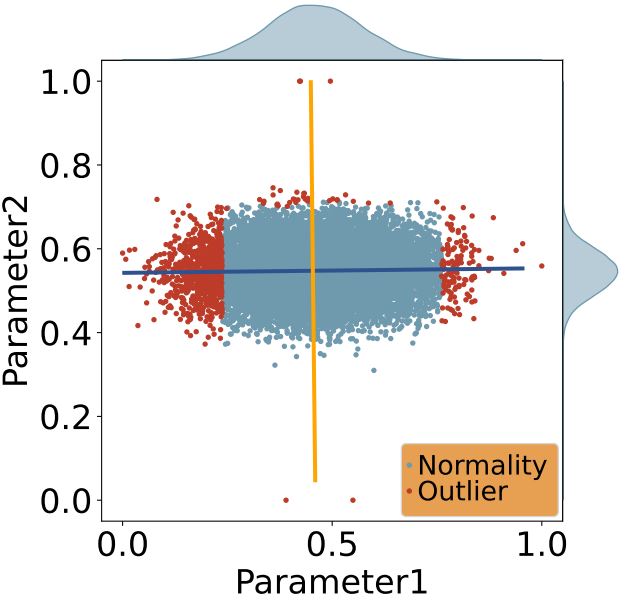}
			\put(-1,80){\large\textbf{c}}
	\end{overpic}}
	\subfigure{
		\label{section1_generate_example2_mse_loss1000}
		\begin{overpic}[width=0.24\textwidth]{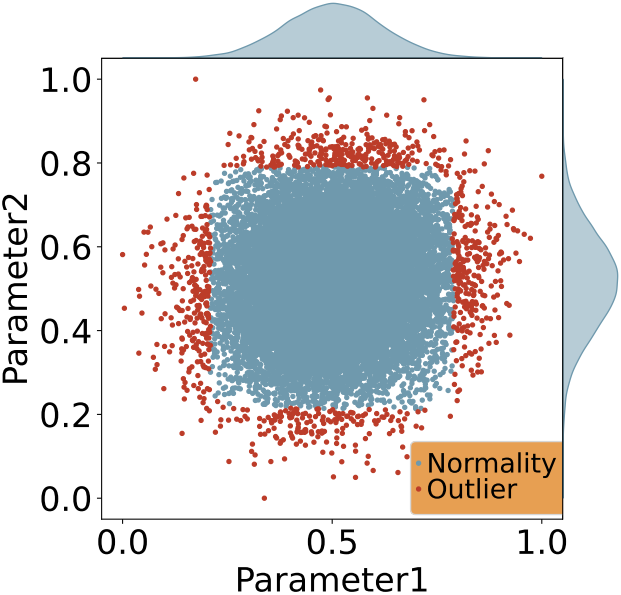}
			\put(-1,80){\large\textbf{d}}
	\end{overpic}}
	\subfigure{
		\label{section1_generate_example3_mse_loss1000}
		\begin{overpic}[width=0.24\textwidth]{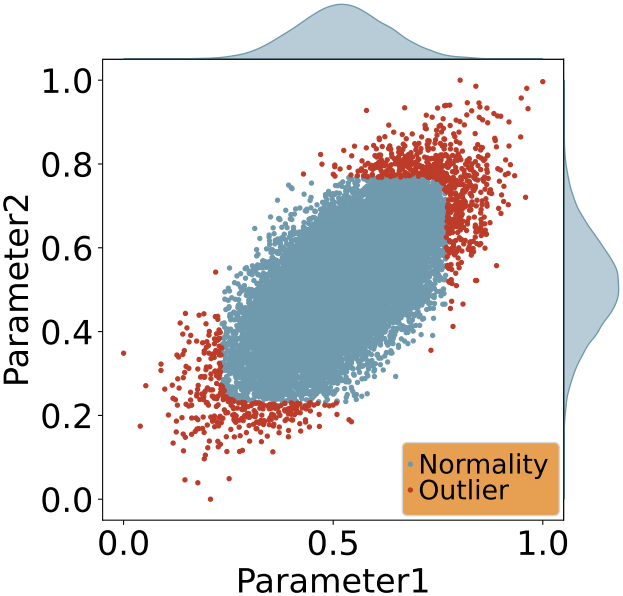}
			\put(-1,80){\large\textbf{e}}
	\end{overpic}}
	
	\subfigure{
		\label{section1_generate_example1_mse_eig_loss_1000}
		\begin{overpic}[width=0.24\textwidth]{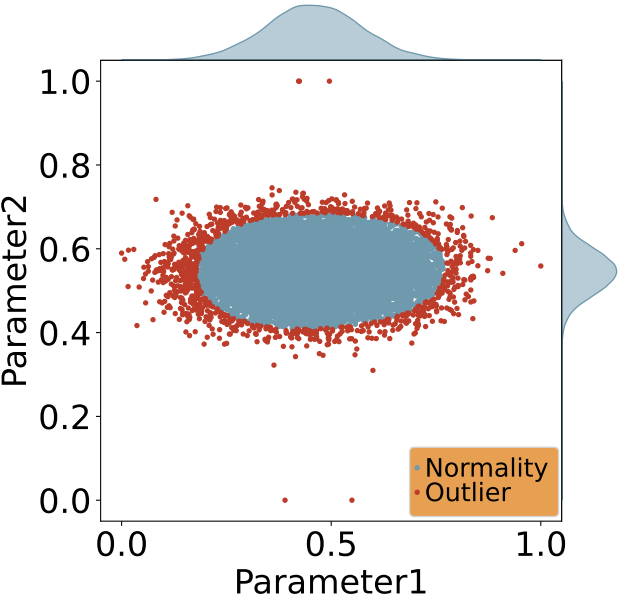}
			\put(-1,80){\large\textbf{f}}
	\end{overpic}}
	\subfigure{
		\label{section1_generate_example2_mse_eig_loss1000}
		\begin{overpic}[width=0.24\textwidth]{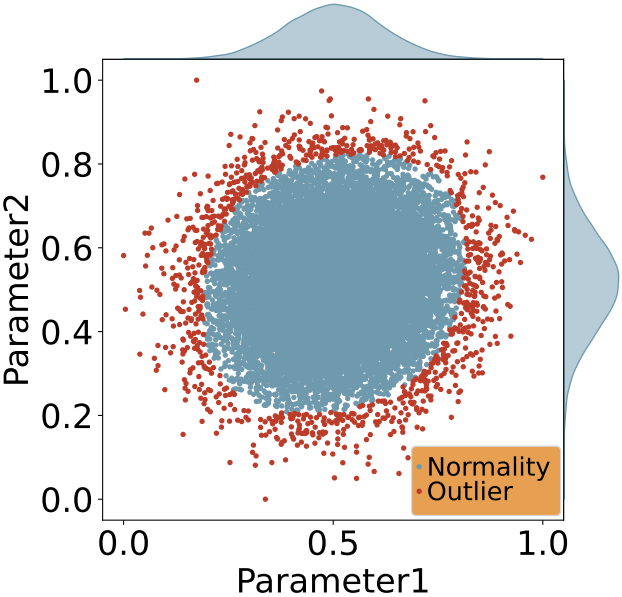}
			\put(-1,80){\large\textbf{g}}
	\end{overpic}}
	\subfigure{
		\label{section1_generate_example3_mse_eig_loss1000}
		\begin{overpic}[width=0.24\textwidth]{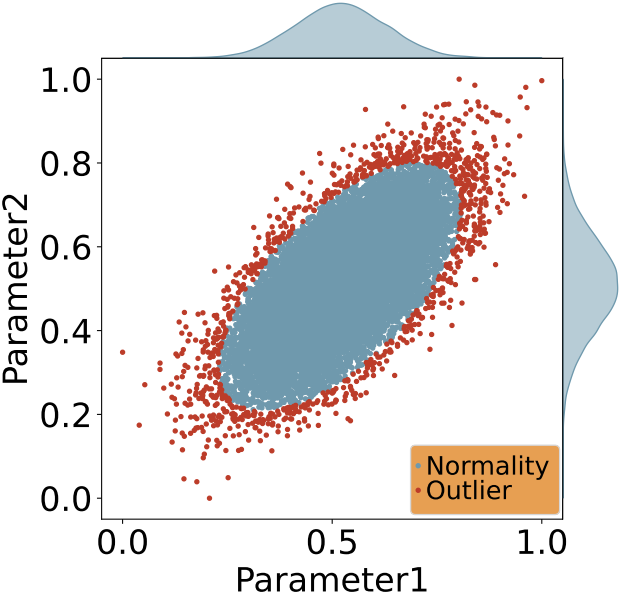}
			\put(-1,80){\large\textbf{h}}
	\end{overpic}}
	
	\subfigure{
		\label{section1_generate_example4_mse_loss1000}
		\begin{overpic}[width=0.3\textwidth]{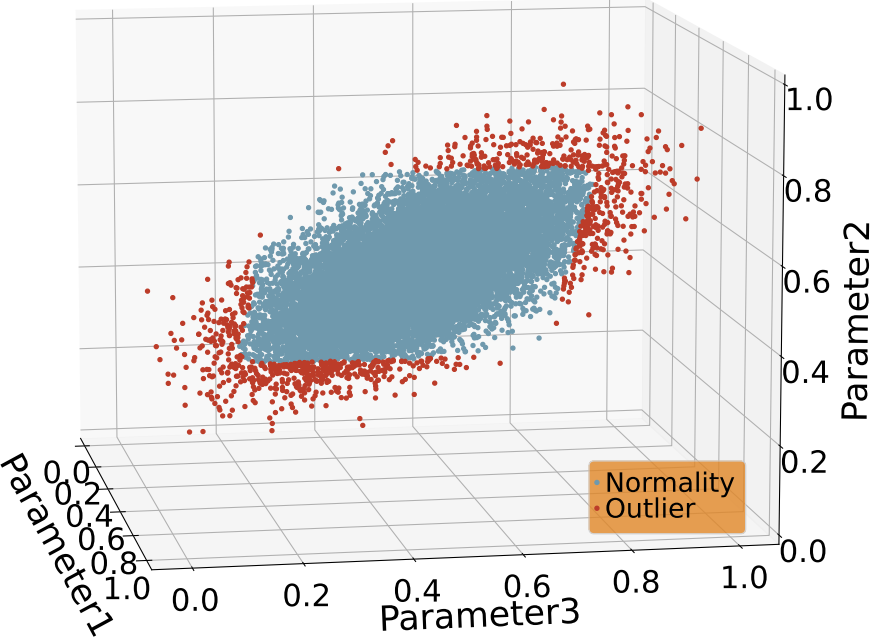}
			\put(1,68){\large\textbf{i}}
	\end{overpic}}
	\hspace{0.8in}
	\subfigure{
		\label{section1_generate_example4_mse_eig_loss1000}
		\begin{overpic}[width=0.3\textwidth]{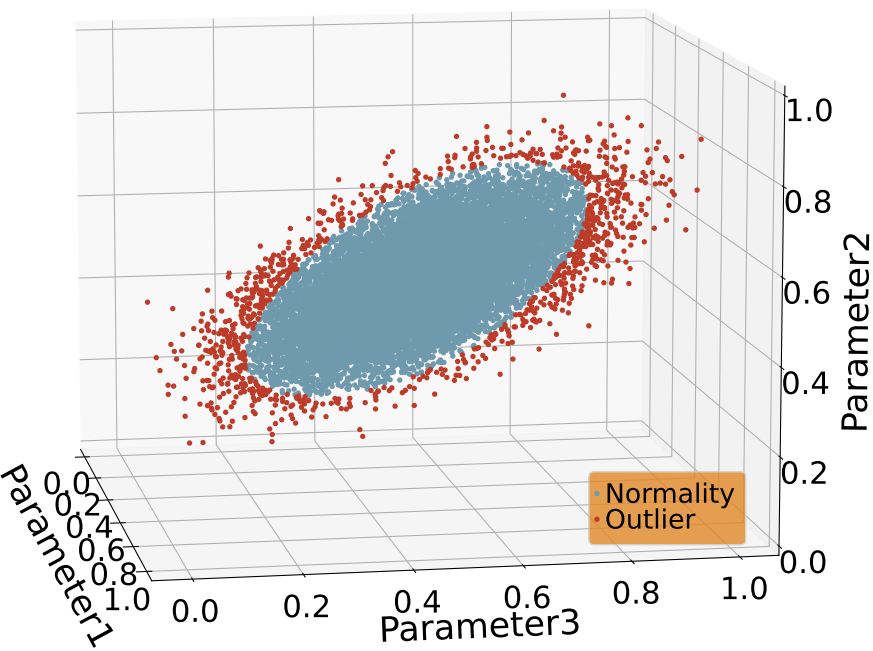}
			\put(1,68){\large\textbf{j}}
	\end{overpic}}
	\caption{{\bf ODBAE accurately identify outliers with multi-dimensional joint anomalies.}  {\bf a}, Scatter plot of parameters BL and BW for mouse developmental dataset, the data points corresponding to mice with the gene \textit{Ckb} knockout deviate significantly from most of the data points, but their values of both BL and BW are within the normal range. The BL or BW values of the data points in the shaded part are ranked in the top $2$\% based on the absolute value of the z-score. {\bf b}, The distribution of standardized BMI of all knockout mice in mouse developmental dataset, the mean BMI of \textit{Ckb} knockout mice is significantly smaller. {\bf c}-{\bf h}, Outlier detection results for $3$ synthetic $2$-dimensional Gaussian distribution datasets when the outlier ratio is $0.05$, including detection results of MSE-trained autoencoder and ODBAE for $2$-dimensional Gaussian distribution dataset with diagonal covariance matrix and non-Gaussian distributed noise ({\bf c}, {\bf f}), $2$-dimensional Gaussian distribution dataset with diagonal covariance matrix ({\bf d}, {\bf g}) and $2$-dimensional Gaussian distribution dataset with non-diagonal covariance matrix ({\bf e}, {\bf h}). The orange line and grey line in ({\bf c}) represent two principal directions. {\bf i}-{\bf j}, Outlier detection results of MSE-trained autoencoder ({\bf i}) and ODBAE ({\bf j}) for $3$-dimensional dataset with an intrinsic dimension of $2$ when the outlier ratio is $0.05$, and the data points follow a 2-dimensional Gaussian distribution with respect to Parameter1 and Parameter3.}
	\label{fig2}
\end{figure}

To illustrate, we analyzed eight developmental parameters: Body length (BL), Body weight (BW), Bone Area (BA), Bone Mineral Density (BMD, excluding skull), Distance travelled total (DT), Forelimb and hindlimb grip strength measurement mean (FHG), heart rate (HR) and Heart weight (HW). Using wild-type mice as the training set and knockout mice as the test set, we evaluated 1904 single-gene knockout mouse strains. Given the prevalence of sexual dimorphism in disease phenotypes\cite{Karp2017PrevalenceOS,Rozman2018IdentificationOG}, males and females were analyzed separately. In the female dataset, ODBAE identified \textit{Ckb} null mice as outliers despite their individual parameter values being within the normal range. These mice had normal BL and BW, but their body mass index (BMI) was abnormally low. Specifically, four of the eight \textit{Ckb} null mice had extremely low BMI values, calculated as $BMI$ $=$ $BW$ (g) $/$ ${BL}^2$ (cm)\cite{Novelli2007AnthropometricalPA,Hu2015AGA,Macdo2020MurinometricMA}. Analysis revealed that while BL and BW were within the expected ranges, their relationship was abnormal, leading to the identification of these mice as outliers (Fig.\ref{fig2}a). The average BMI of these \textit{Ckb}-deficient mice was lower than 97.14\% of other mice (Fig.\ref{fig2}b), demonstrating the sensitivity of ODBAE to complex, multivariate outliers. In addition, \textit{Ckb} has previously been implicated in developmental processes and obesity, further strengthening the biological significance of these findings\cite{Rahbani2021CreatineKB}.

ODBAE also identified previously unknown phenotypes, highlighting its potential to uncover novel gene functions. In terms of the identification of HLP, most studies choose MSE as the training loss function and force the autoencoder to completely reconstruct the training dataset, which brings two major limitations: (1) only HLP along principal component directions with poor reconstruction can be detected; (2) in the process of anomaly detection, each dimension of the dataset is independently detected. To evaluate the performance difference between our modified loss function and MSE in HLP detection, we used these two trained autoencoder to detect outliers for several datasets following multi-dimensional Gaussian distributions. For example, data in Fig.\ref{fig2}c-\ref{fig2}e were three 2-dimensional Gaussian distribution datasets and data in Fig.\ref{fig2}i was 3-dimensional dataset whose intrinsic dimension was two, the data points followed a 2-dimensional Gaussian distribution with respect to Parameter1 and Parameter3. The detection result in Fig.\ref{fig2}c shows that most of the detected HLP lie in the direction of the principal components indicated by the blue line but few lie in the other direction. In Fig.\ref{fig2}d, HLP in two dimensions of the dataset are independently detected, making outliers that are anomalous in multiple dimensions (outliers at the four corners) undetectable. Fig.\ref{fig2}i demonstrates the same issue in three-dimensional space. Moreover, these issues persisted when the activation function was replaced with another function (Supplementary Fig.2,3). This is because the saturating nature of the non-linear activation function causes the autoencoder to have different reconstruction capabilities for data points located at the boundaries of the dataset compared to those located in the interior (Methods, Supplementary Fig.4a, 4c). Usually, the distribution of a dataset along various principal component directions may not be balanced, and the number of data points falling into unreconstructable regions may also be uneven (Methods). This leads to different detection performances for HLP along different principal component directions. Besides, the distribution of reconstructed dataset is independent across all dimensions (Supplementary Fig.4a, 4c). Therefore, the new loss function in ODBAE is designed to ensure that the reconstructed dataset follows a balanced joint distribution (Supplementary Fig.4b, 4d). Then, we used ODBAE to detect outliers in these datasets and found that most outliers were identified (Fig.\ref{fig2}f-\ref{fig2}h, \ref{fig2}j).

We further evaluated ODBAE's detection of both IP and HLP using a synthetic 3-dimensional dataset (Supplementary Fig. 5a). Area Under Curve (AUC) and Average Precision (AP) scores were used to evaluate the accuracy of outlier detection results (Methods). The generated 3-dimensional training dataset formed a 2-dimensional manifold, where the subspace represented by Parameter1 and Parameter3 followed a 2-dimensional Gaussian distribution (Supplementary Fig.5a). Here, we assumed that the ratio of HLP and the ratio of IP were equal. During the anomaly detection, we used the training dataset as the test dataset. Then, we considered the top $\lfloor \delta n \rfloor$ data points in the subspace represented by Parameter1 and Parameter3, ranked by Mahalanobis distance, as HLP, where $\delta$ was the outlier ratio, and $n$ was the number of the data points in the training dataset. Besides, we additionally generated $\lfloor \delta n \rfloor$ points that were not on the manifold of the training dataset and treated them as IP (Supplementary Fig.5b). The results showed that ODBAE worked better than other anomaly detection methods (Supplementary Fig.5c,5d).

In summary, ODBAE is a powerful tool for detecting complex phenotypes in high-dimensional biological datasets, capturing both isolated and coordinated abnormalities. This capability allows the identification of novel gene functions and previously undetected phenotypes, particularly in cases where traditional methods fail to detect subtle perturbations in physiological homeostasis.

\subsection*{ODBAE outperforms other methods by a large margin}
To evaluate the strengths of ODBAE, we performed outlier detection of ODBAE on low-dimensional synthetic datasets, high-dimensional synthetic datasets and two benchmark datasets. ODBAE was compared with the autoencoder with MSE loss function (MSE-AE), autoencoder with Mean Absolute Error loss function (MAE-AE) and Deep Autoencoding Gaussian Mixture Model (DAGMM) \cite{Zong2018DeepAG}. The comparison results were presented through AUC and AP scores.

\begin{figure}[htbp]
	\centering
	\hspace{-0.4in}
	\subfigure{
		\label{section4_generate_example1_data}
		\begin{overpic}[width=0.19\textwidth]{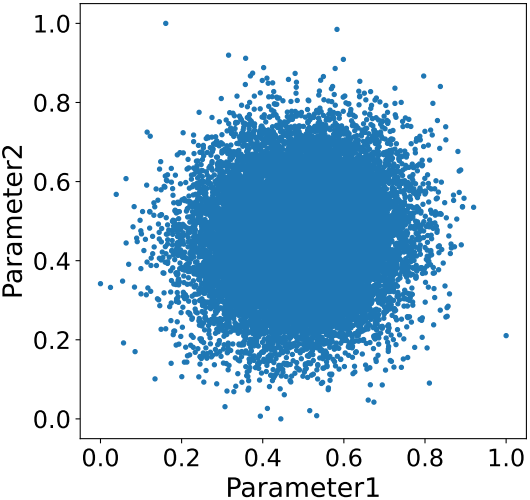}
			\put(2,95){\large\textbf{a}}
	\end{overpic}}
	\hspace{0.3in}
	\subfigure{
		\label{section4_generate_example2_data}
		\begin{overpic}[width=0.19\textwidth]{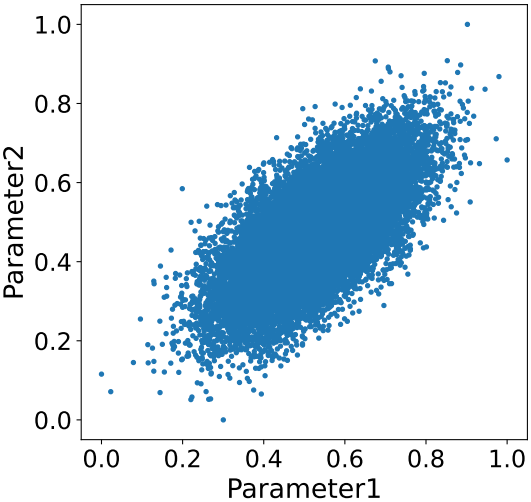}
			\put(2,95){\large\textbf{b}}
	\end{overpic}}
	\hspace{0.3in}
	\subfigure{
		\label{section4_generate_example3_data}
		\begin{overpic}[width=0.19\textwidth]{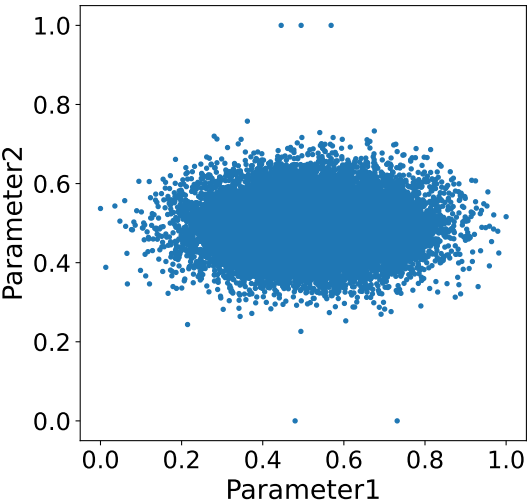}
			\put(2,95){\large\textbf{c}}
	\end{overpic}}
	
	\subfigure{
		\label{section4_gnerate_example1_auc_score}
		\begin{overpic}[width=0.24\textwidth]{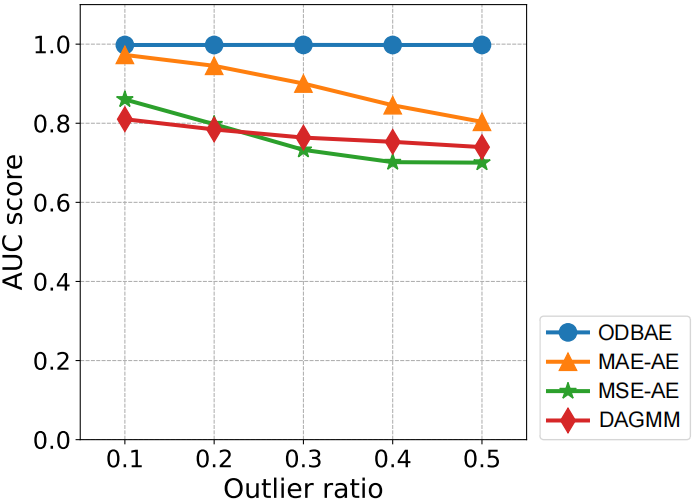}
			\put(2,70){\large\textbf{d}}
	\end{overpic}}
	\subfigure{
		\label{section4_gnerate_example2_auc_score}
		\begin{overpic}[width=0.24\textwidth]{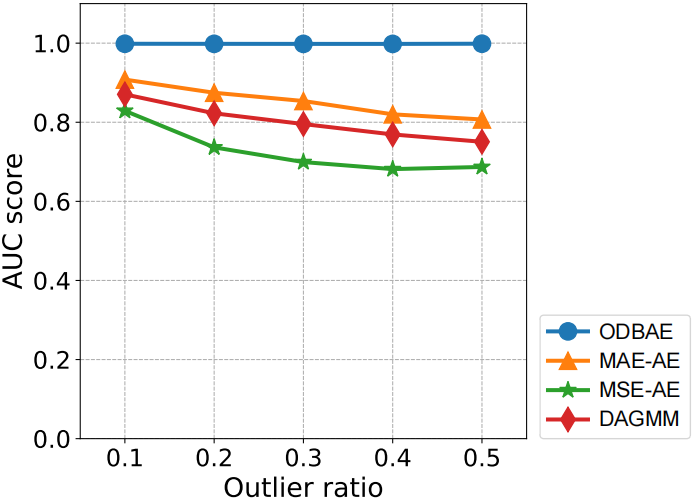}
			\put(2,70){\large\textbf{e}}
	\end{overpic}}
	\subfigure{
		\label{section4_gnerate_example3_auc_score}
		\begin{overpic}[width=0.24\textwidth]{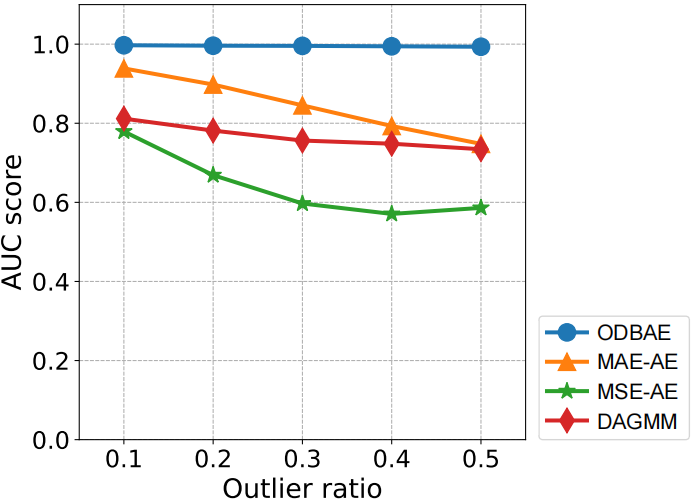}
			\put(2,70){\large\textbf{f}}
	\end{overpic}}

	\subfigure{
		\label{section4_gnerate_example1_ap_score}
		\begin{overpic}[width=0.24\textwidth]{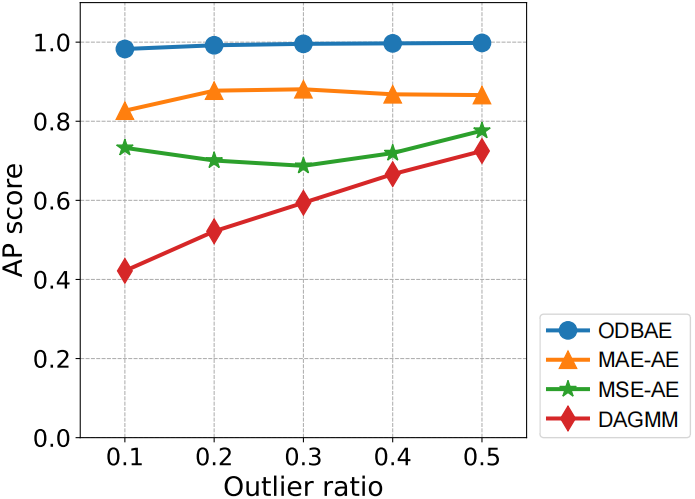}
			\put(2,70){\large\textbf{g}}
	\end{overpic}}
	\subfigure{
		\label{section4_gnerate_example2_ap_score}
		\begin{overpic}[width=0.24\textwidth]{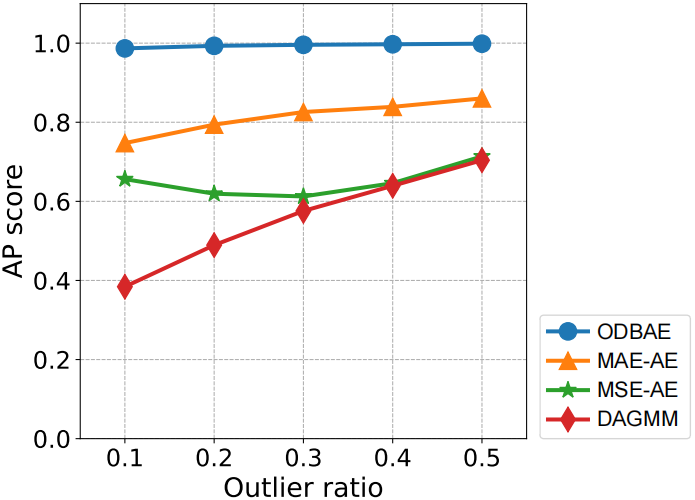}
			\put(2,70){\large\textbf{h}}
	\end{overpic}}
	\subfigure{
		\label{section4_gnerate_example3_ap_score}
		\begin{overpic}[width=0.24\textwidth]{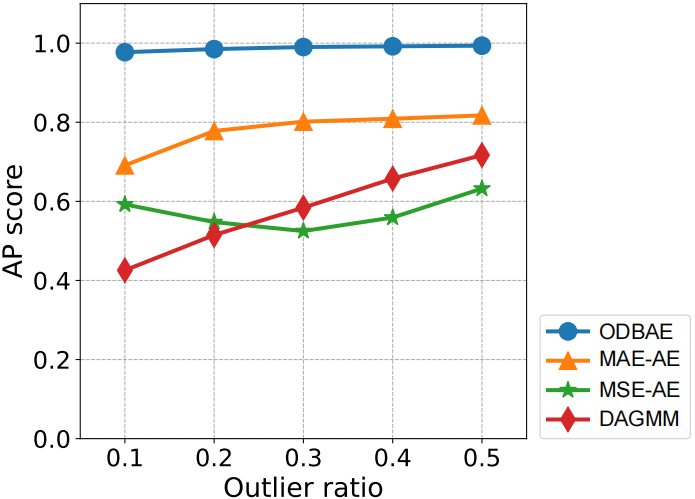}
			\put(2,70){\large\textbf{i}}
	\end{overpic}}
	
	\subfigure{
		\label{section4_dry_bean_auc_score}
		\begin{overpic}[width=0.38\textwidth]{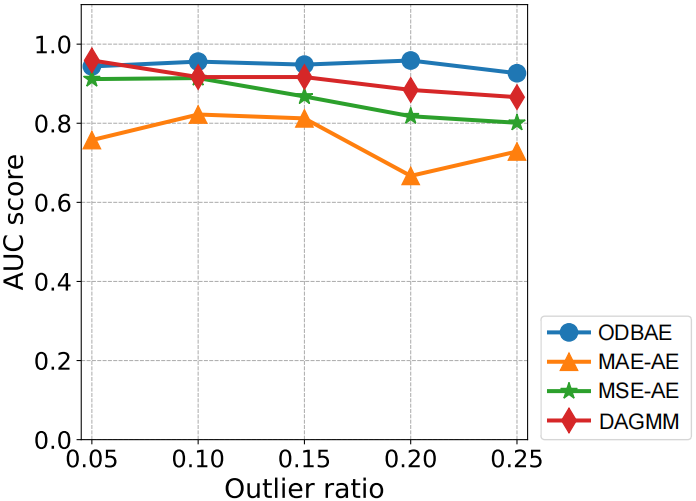}
			\put(2,70){\large\textbf{j}}
	\end{overpic}}
	\subfigure{
		\label{section4_dry_bean_ap_score}
		\begin{overpic}[width=0.38\textwidth]{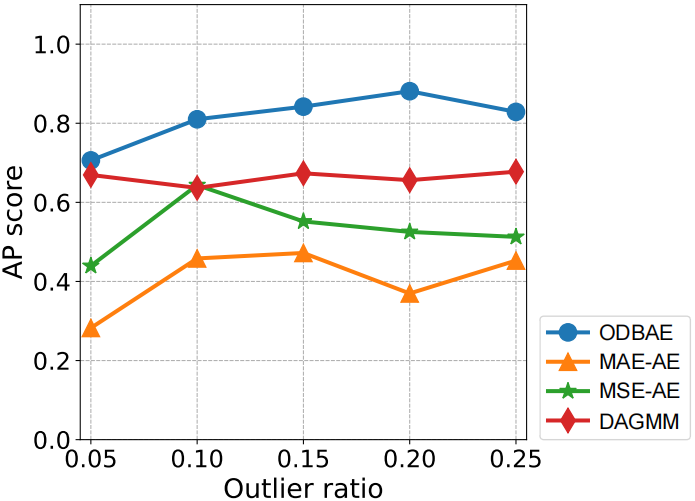}
			\put(2,70){\large\textbf{k}}
	\end{overpic}}

	\subfigure{
		\label{section4_cancer_auc_score}
		\begin{overpic}[width=0.38\textwidth]{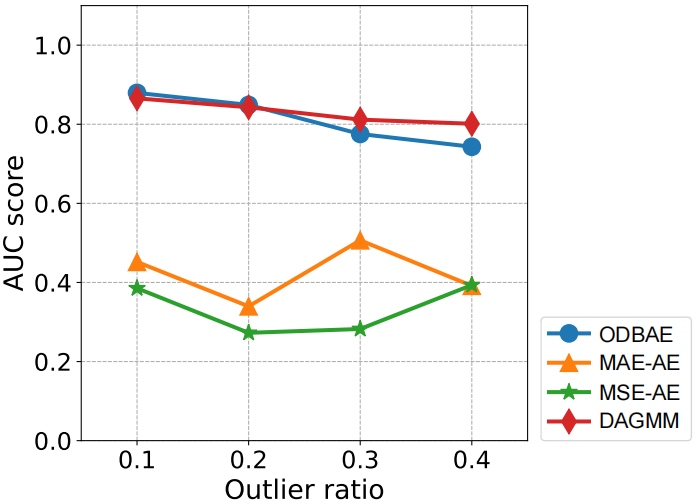}
			\put(2,70){\large\textbf{l}}
	\end{overpic}}
	\subfigure{
		\label{section4_cancer_ap_score}
		\begin{overpic}[width=0.38\textwidth]{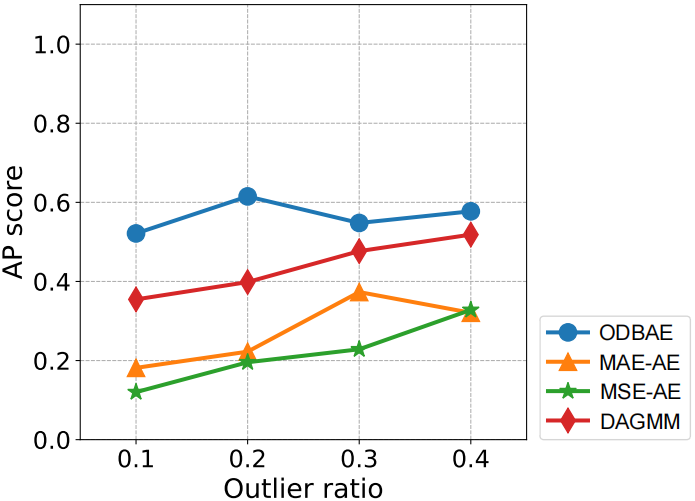}
			\put(2,70){\large\textbf{m}}
	\end{overpic}}
	\caption{{\bf Outlier detection effect of ODBAE is generally better than other schemes.} {\bf a}-{\bf c}, Data point distribution plot for $3$ synthetic dataset, including $2$-dimensional Gaussian distribution dataset with diagonal covariance matrix ({\bf a}), $2$-dimensional Gaussian distribution dataset with non-diagonal covariance matrix ({\bf b}) and $2$-dimensional Gaussian distribution dataset with diagonal covariance matrix and non-Gaussian distributed noise ({\bf c}). {\bf d}-{\bf f}, AUC scores of comparison results of dataset in Fig.\ref{fig3}a ({\bf d}), Fig.\ref{fig3}b ({\bf e}) and Fig.\ref{fig3}c ({\bf f}). {\bf g}-{\bf i}, AP scores of comparison results of dataset in Fig.\ref{fig3}a ({\bf g}), Fig.\ref{fig3}b ({\bf h}) and Fig.\ref{fig3}c ({\bf i}). {\bf j},{\bf k}, Comparison results for Dry Bean Dataset. {\bf l},{\bf m}, Comparison results for Breast Cancer Dataset.}
	\label{fig3}
\end{figure}

We generated three datasets with $2$-dimensional Gaussian distribution, and their intrinsic dimension were also $2$. In one of the datasets, the correlation between the two dimensions was small (Fig.\ref{fig3}a); the correlation between the two dimensions in another dataset was relatively large (Fig.\ref{fig3}b); in the last dataset, the correlation between the two dimensions was also small and non-Gaussian distributed noise existed (Fig.\ref{fig3}c). Therefore, the outliers in these three datasets were all HLP. Then we set different outlier ratios and calculated the corresponding AUC and AP scores of the outlier detection results. To be specific, if the outlier ratio is $\delta$ and the number of sample points is $n$, then the input data is sorted according to their Mahalanobis distance, and the first $\lfloor \delta n \rfloor$ sample points are regarded as positive. Fig.\ref{fig3}d-\ref{fig3}f show the AUC scores of test results on these three datasets, and Fig.\ref{fig3}g-\ref{fig3}i are corresponding AP scores, respectively. ODBAE outperformed other schemes in detecting HLP in all three datasets at any outlier ratio.

Since autoencoders are often used for outlier detection in high-dimensional datasets, we also analyzed the outlier detection results of ODBAE on two high-dimensional synthetic datasets. Both datasets followed multi-dimensional Gaussian distribution and their covariance matrices were both diagonal matrices. So the intrinsic dimension of these two datasets were equal to their actual dimension. One of the datasets was $50$ dimensional and the other dataset was $100$ dimensional. Finally, ODBAE still performed best in HLP detection on high-dimensional datasets (Supplementary Fig.6).

Additionally, we tested ODBAE on two benchmark datasets: Dry Bean Dataset and Breast Cancer Dataset. Dry Bean Dataset contained $13611$ instances and $17$ attributes, and there were $7$ classes in the original dataset. In each experiment, the inliers were $2000$ data points from a certain class and $\delta \times 2000$ data points were randomly selected from the remaining $6$ classes of data points as outliers, where $\delta$ was the outlier ratio and $\delta \in \{0.05,0.1,0.15,0.2,0.25\}$. For this dataset, we set the dimension of the hidden layer of the autoencoder to $8$ during the experiment. Breast Cancer Dataset contained $569$ instances and $31$ attributes. All data points fell into two categories: benign and malignant. We randomly selected $300$ data points from the data points labeled benign as inliers and $\delta \times 300$ data points from the other categories as outliers. For this dataset, the outlier ratio $\delta \in \{0.1,0.2,0.3,0.4\}$ and the dimension of the hidden layer of the autoencoder was $15$. Overall, due to the improvement in HLP detection, ODBAE worked best on both benchmark datasets (Fig.\ref{fig3}j-\ref{fig3}m).

In summary, ODBAE outperforms several existing autoencoder-based methods in detecting anomalies from both human-induced and benchmark datasets, indicating that ODBAE can more comprehensively identify anomalous situations. Therefore, ODBAE provides strong support for finding unknown anomalies and exploring unknown phenotypes in existing biological data.

\begin{figure}[htbp]
	\centering
	\subfigure{
		\label{Male_Female_gene_result}
		\begin{overpic}[width=0.45\textwidth]{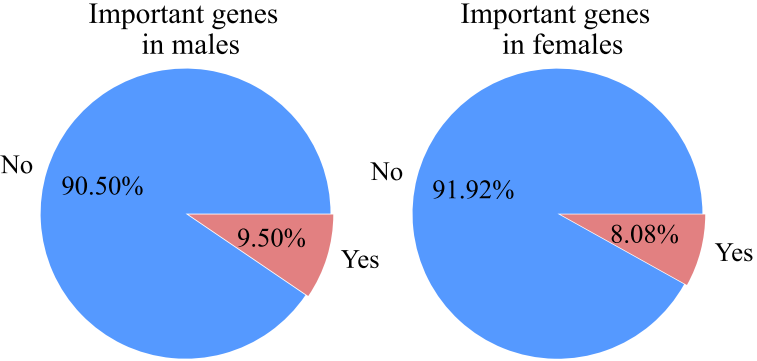}
			\put(-1,40){\large\textbf{a}}
	\end{overpic}}
	\subfigure{
		\label{section5_abnormal_genes}
		\begin{overpic}[width=0.25\textwidth]{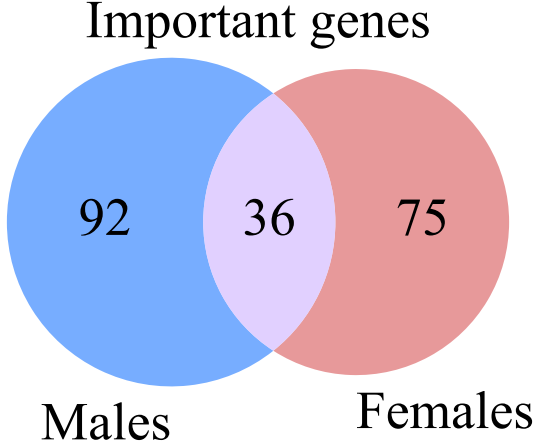}
			\put(-1,70){\large\textbf{b}}	
	\end{overpic}}
	
	\subfigure{
		\label{Alb_HDLC_Crabp1}
		\begin{overpic}[width=0.25\textwidth]{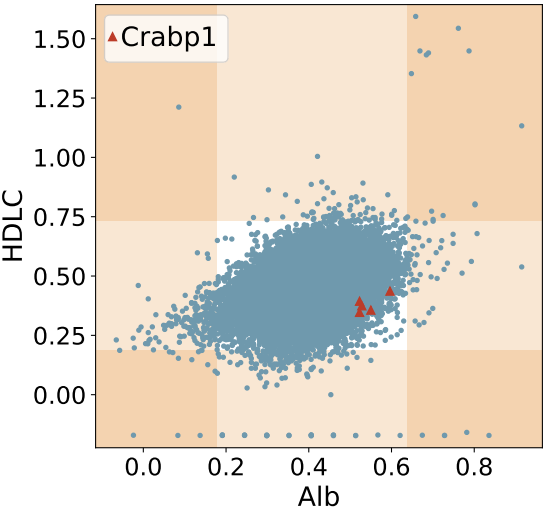}
			\put(1,90){\large\textbf{c}}
	\end{overpic}}
	\hspace{0.1in}
	\subfigure{
		\label{Metabolism-related_in_MGI} 
		\begin{overpic}[width=0.5\textwidth]{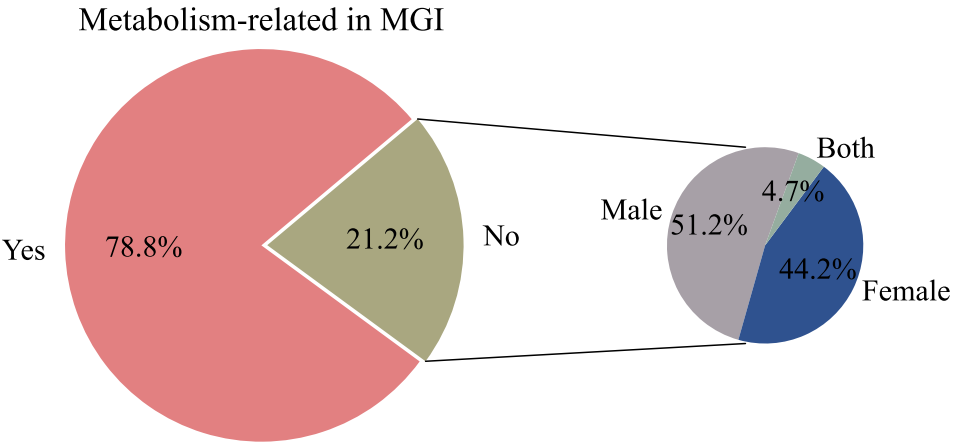}
			\put(-1,45){\large\textbf{d}}
	\end{overpic}}
	
	\subfigure{
		\label{43gene_pathway} 
		\begin{overpic}[width=0.7\textwidth]{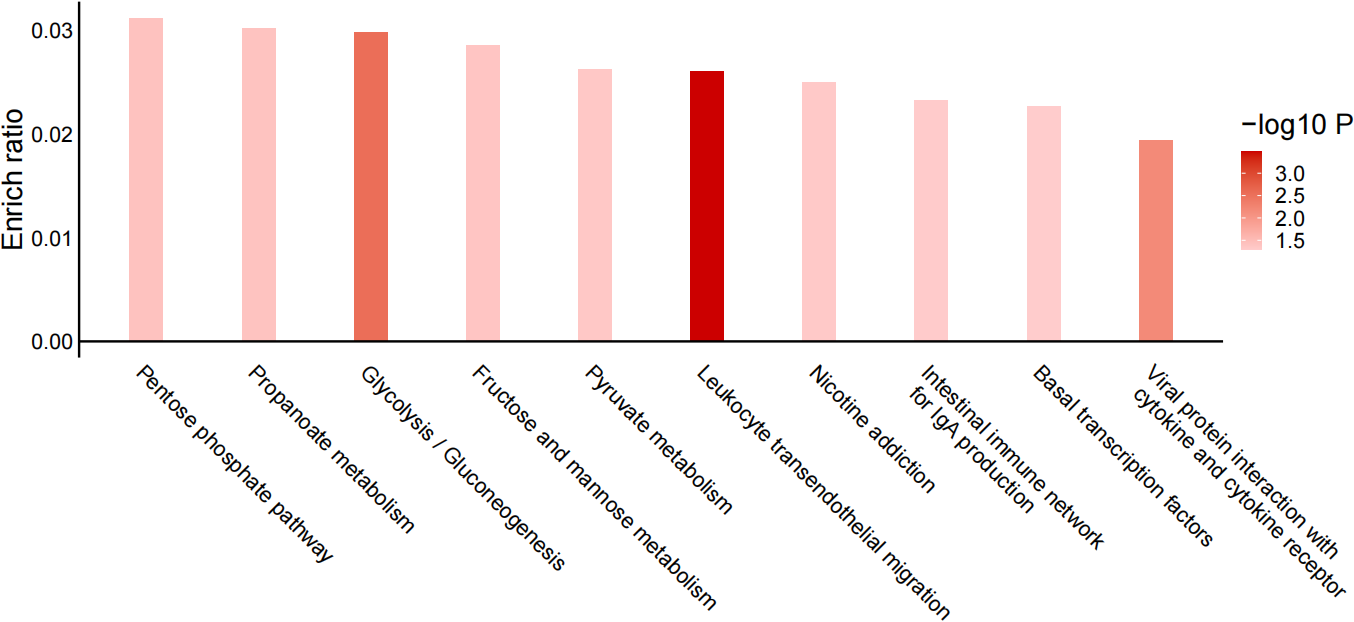}
			\put(-1,46){\large\textbf{e}}
	\end{overpic}}
	
	\subfigure{
		\label{43gene_pathway_huamn} 
		\begin{overpic}[width=0.7\textwidth]{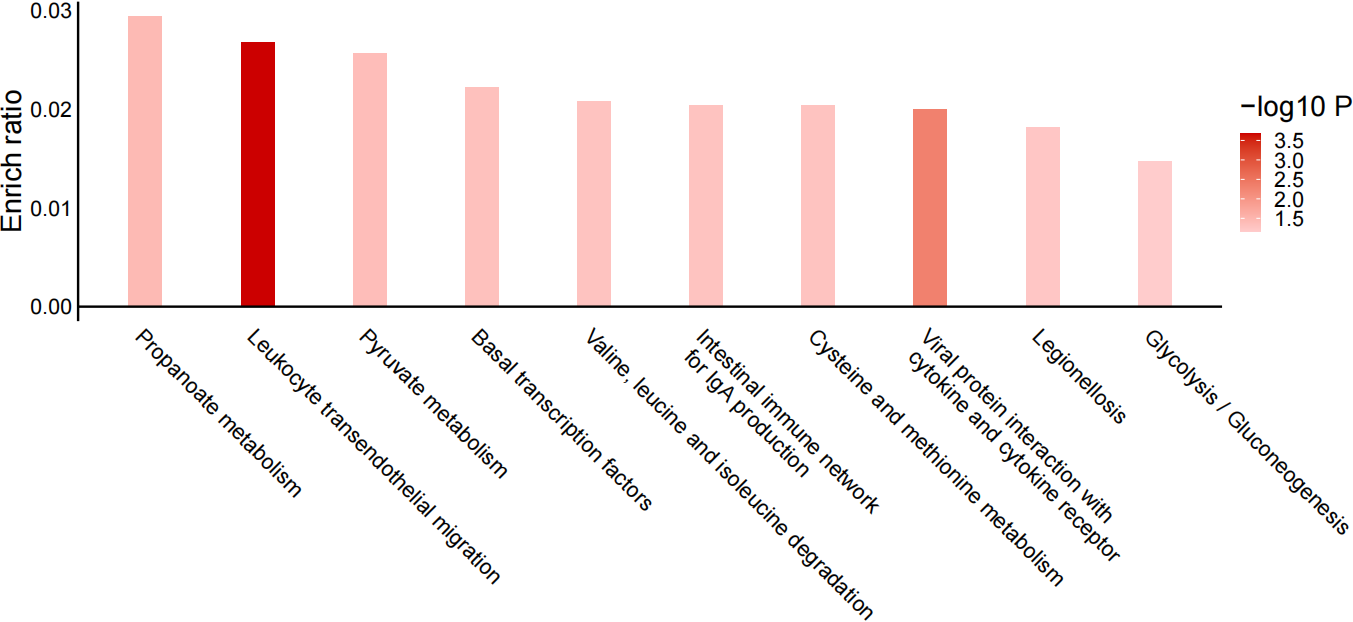}
			\put(-1,46){\large\textbf{f}}
	\end{overpic}}
	\caption{{\bf ODBAE identifies new metabolic genes from a metabolism-related dataset of knockout mice.} {\bf a}, The proportion of important genes identified in the male and female datasets. {\bf b}, ODBAE identified $128$ important genes in the male dataset and $111$ important genes in the female dataset, for a total of $203$ important genes. {\bf c}, Visualization results of outliers corresponding to gene \textit{Crabp1} in the subspace formed by abnormal parameters. The absolute z-score values of Alb or HDLC of the data points in the shaded part are ranked in the top $1.2$\%. {\bf d}, Validation results of important genes in the MGI database. $78.8$\% of the genes are associated with metabolism, and $21.2$\% are newly identified by ODBAE. Besides, $51.2$\% and $44.2$\% new metabolic genes are identified in the male or female datasets, respectively, and $4.7$\% are detected in both male and female datasets. {\bf e},{\bf f}, Histogram of KEGG pathway enrichment for $43$ novel genes in mice ({\bf e}) and humans ({\bf f}) obtained through KOBAS.}
	\label{fig4}
\end{figure}

\subsection*{ODBAE identifies new metabolic genes}
We applied ODBAE to metabolism-related datasets from IMPC to uncover genetic elements involved in metabolic regulation. Fourteen key metabolic parameters were selected for analysis, including: Alanine aminotransferase (ALA), Albumin (Alb), Alkaline phosphatase (ALP), Aspartate aminotransferase (ASA), Calcium (Ca), Creatinine (Cre), Glucose (Glu), HDL-cholesterol (HDLC), Phosphorus (Ph), Total bilirubin (TB), Total cholesterol (TC), Total protein (TP), Triglycerides (TG), and Urea. In total, we analyzed 45922 mice from 3064 single-gene knockout strains, including 23024 males and 22898 females.

ODBAE was used to detect outliers in both male and female data sets. Initially, mice with absolute z-scores in the top 1.2\% for at least one metabolic parameter were flagged as outliers. Of the 23024 male mice, 2672 were flagged as abnormal, and of the 22898 female mice, 2553 abnormalities were identified. To ensure robust detection, we adjusted the proportion of anomalies and identified the top 10\% of mice with the highest reconstruction errors as outliers. Genes were considered significant if more than 50\% of mice from a given knockout strain were identified as outliers, suggesting a strong association between these genes and metabolism. In total, ODBAE identified 128 significant genes in males (9.5\%) and 111 significant genes in females (8.08\%), for a total of 203 genes (Fig.\ref{fig4}a, \ref{fig4}b)

Notably, 91 of these 203 genes could not be detected using traditional z-score based methods. This indicates that the metabolic abnormalities associated with these genes involve complex, multi-parameter phenotypes. ODBAE's anomaly detection includes a detailed explanation of outliers based on the highest reconstruction error and kernel SHAP analysis, which allows us to identify the specific metabolic parameters driving the anomalies. For example, knockout of the \textit{Crabp1} gene resulted in abnormal levels of Alb and HDLC, even though the individual z-scores for these parameters were within normal ranges. Visualization of these data (Fig.\ref{fig4}c) shows that while the z-scores for Alb and HDLC appeared normal, the knockout mice deviated significantly from the central distribution of all data points, highlighting the complex metabolic perturbation caused by \textit{Crabp1} knockout. More examples are shown in Supplementary Fig.7.

Further investigation using the Mouse Genome Informatics (MGI) database revealed that 43 of the 203 genes (21.18\%) identified by ODBAE had no previously known association with metabolic phenotypes (Fig.\ref{fig4}d). To explore the potential relevance of these newly identified genes, we analyzed their human orthologues. Single-nucleotide polymorphisms (SNPs) within a $\pm$ 1 kb region of these orthologues were extracted from GWAS Central\cite{Beck2022GWASCA}, resulting in a total of 804 SNPs. We then investigated their association with metabolic diseases, focusing on type 2 diabetes (T2D). Specifically, for each SNP, we evaluated the extent of association across T2D-related traits on data from the DIAGRAM, GIANT and GLGC consortia\cite{MorrisVoightTeslovichFerreiraSegrSteinthorsd2,Ried2016APC,Willer2013DiscoveryAR}. Cross-phenotype meta-analysis (CPMA)\cite{Cotsapas2011PervasiveSO} identified SNPs in four gene regions—\textit{TWF2}, \textit{TMED10}, \textit{HOXA10}, and \textit{NBAS}—that were strongly linked to T2D-related traits (CPMA $p<0.05$). Visualization of the corresponding outliers confirmed that these genes exhibited distinct patterns of deviation across key metabolic dimensions (Supplementary Fig.8). In addition to T2D, many of these novel genes may be associated with other metabolic disorders. KEGG pathway analysis (via KOBAS) for both mouse and human datasets revealed significant enrichment in metabolic pathways\cite{Bu2021KOBASiIP}, including glycolysis, propanoate metabolism, and pyruvate metabolism (Fig.\ref{fig4}e,\ref{fig4}f). These results suggest that ODBAE not only identifies genes involved in complex metabolic phenotypes but also pinpoints pathways that are disrupted by genetic perturbations.

Overall, ODBAE provides a powerful tool for uncovering novel metabolic genes and phenotypes that are undetectable using traditional z-score-based methods. It successfully identifies genes associated with complex, multi-parameter metabolic abnormalities and reveals novel genetic contributors to metabolic diseases, many of which were previously unannotated in the MGI database.

\subsection*{ODBAE integrates novel metabolic phenotypes}
ODBAE's ability to detect outliers based on joint abnormalities, beyond traditional z-score methods, highlights its strength in integrating complex metabolic phenotypes. Here, we demonstrate how ODBAE can detect metabolic abnormalities in gene knockout mice by identifying abnormal relationships between multiple metabolic parameters, rather than focusing on single-parameter outliers. Our analysis of metabolism-related datasets from IMPC reveals that a large proportion of outliers (89.14\% of male and 92.75\% of female mice) exhibit abnormalities in multiple metabolic parameters simultaneously. These findings suggest that joint abnormalities, or perturbations in the correlations between parameters, are important for detecting phenotypic changes in knockout mice. For example, knockout of the \textit{Ckb} gene disrupts the relationship between BW and BL, resulting in an abnormal BMI, even though BW and BL are individually within normal ranges. This highlights how gene knockouts often disrupt intrinsic correlations between physiological parameters, resulting in complex metabolic phenotypes.

\begin{figure}[htbp]
	\centering
	\subfigure{
		\label{male_parameter_relation_new}
		\begin{overpic}[width=0.35\textwidth]{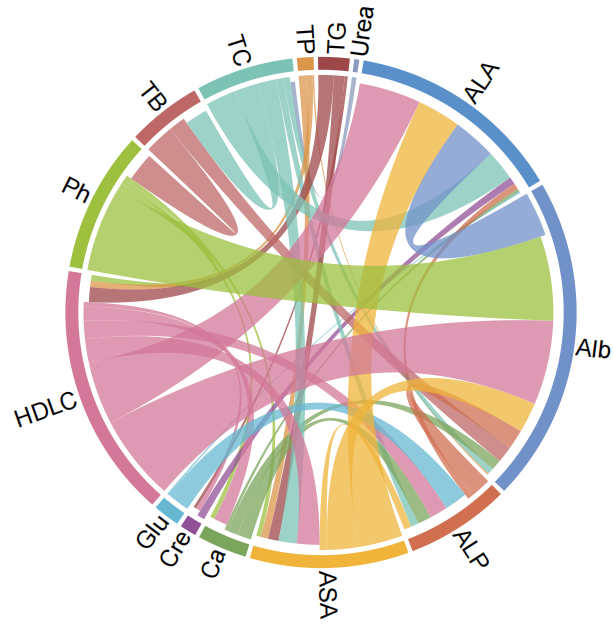}
			\put(-2,75){\large\textbf{a}}
	\end{overpic}}
	\subfigure{
		\label{female_parameter_relation_new}
		\begin{overpic}[width=0.35\textwidth]{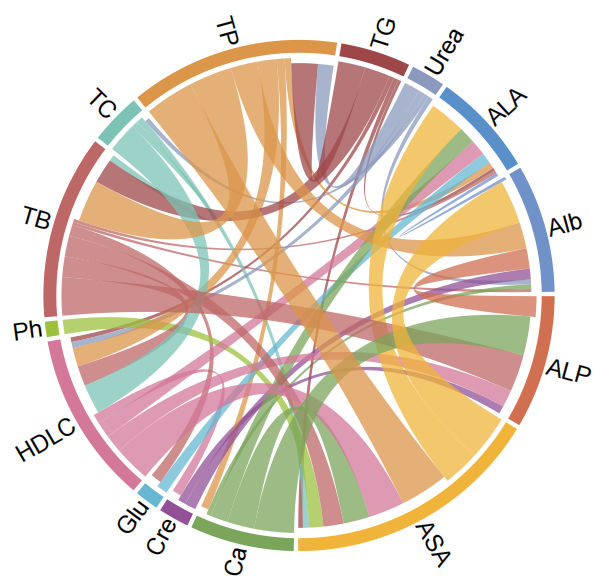}
			\put(-2,75){\large\textbf{b}}
	\end{overpic}}
	
	\subfigure{
		\label{abnormal_parameter_pairs}
		\begin{overpic}[width=0.7\textwidth]{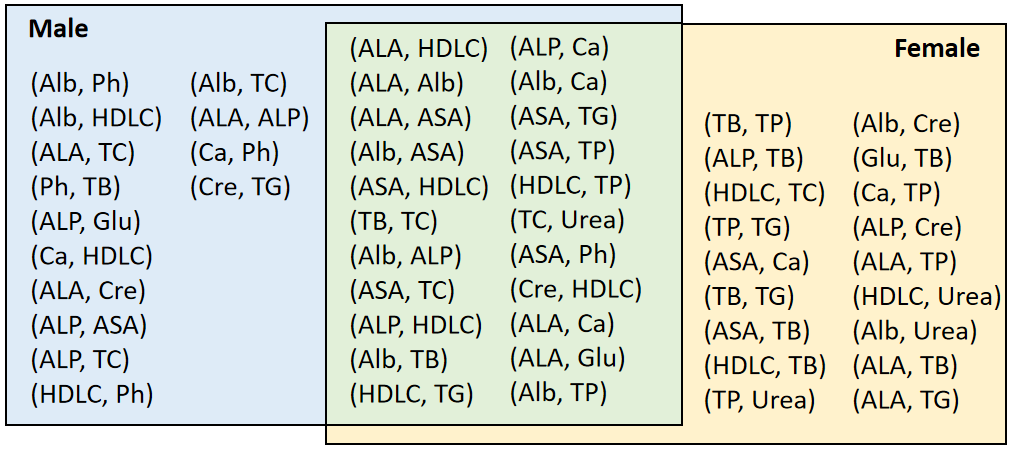}
			\put(-2,45){\large\textbf{c}}
	\end{overpic}}

	\subfigure{
		\label{append_fig1_male}
		\begin{overpic}[width=0.35\textwidth]{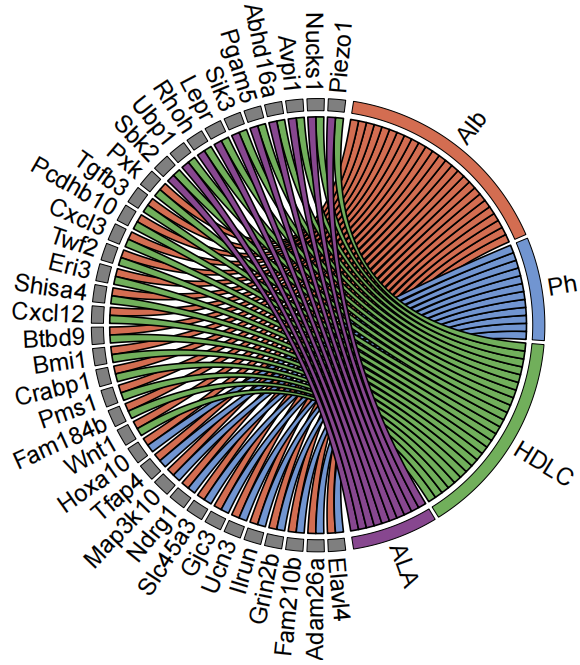}
			\put(-2,80){\large\textbf{d}}
	\end{overpic}}
	\subfigure{
		\label{append_fig2_female}
		\begin{overpic}[width=0.35\textwidth]{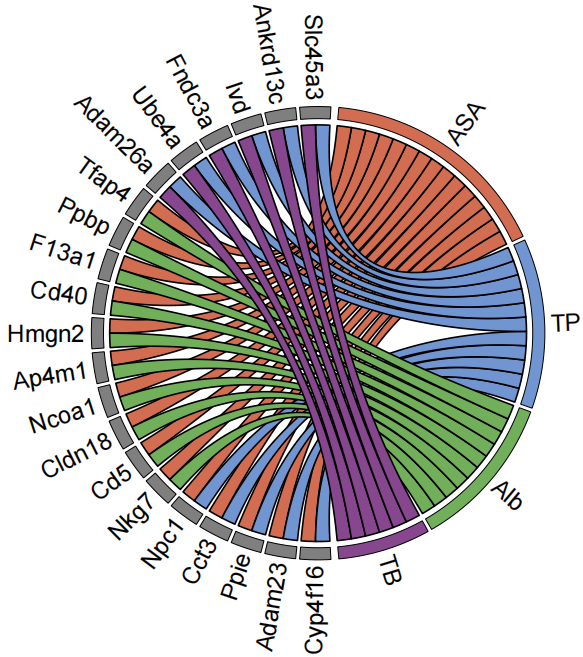}
			\put(-2,85){\large\textbf{e}}
	\end{overpic}}
	\caption{{\bf ODBAE integrates metabolic parameter pairs tend to simultaneous abnormalities.} {\bf a},{\bf b}, Chord diagram plotting the inter-connectivity of metabolic parameters for males (a) and females (b). The outer segments represent the metabolic parameters, the size of the arcs connecting the perameters is proportional to the number of outliers associated. {\bf c}, Abnormal parameter pairs observed in males and females. {\bf d}, {\bf e}, Links between genes and simultaneous abnormal metabolic parameters in males (d) and females (e).}
	\label{fig5}
\end{figure}

To further explore these correlations, we identified parameter pairs most likely to show joint abnormalities in both male and female mice. In males, 36 different abnormal parameter pairs were observed, with the most common being (Alb, Ph), (Alb, HDLC) and (ALA, HDLC) (Fig.\ref{fig5}a,\ref{fig5}c). In females, 40 different abnormal parameter pairs were identified, with the most common being (ASA, TP), (Alb, ASA) and (TB, TP) (Fig.\ref{fig5}b,\ref{fig5}c). Their corresponding knockout genes are shown in Fig.\ref{fig5}d,\ref{fig5}e. Notably, 22 abnormal parameter pairs were found in both males and females, with the most common pairs being (ALA, ASA), (ALA, HDLC) and (Alb, ASA). These common abnormalities suggest conserved metabolic pathways that are affected by specific gene knockouts in both sexes.

\begin{figure}[htbp]
	\centering
	\subfigure{
		\label{ALA_HDLC_Lepr}
		\begin{overpic}[width=0.26\textwidth]{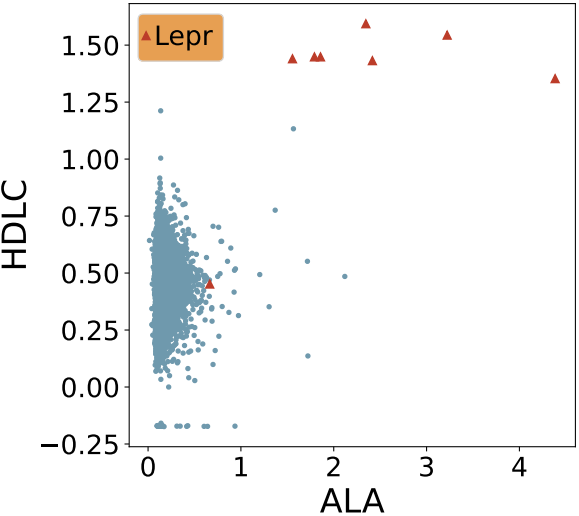}
			\put(-2,88){\large\textbf{a}}
	\end{overpic}}
	\subfigure{
		\label{ALA_HDLC_ratio}
		\begin{overpic}[width=0.385\textwidth]{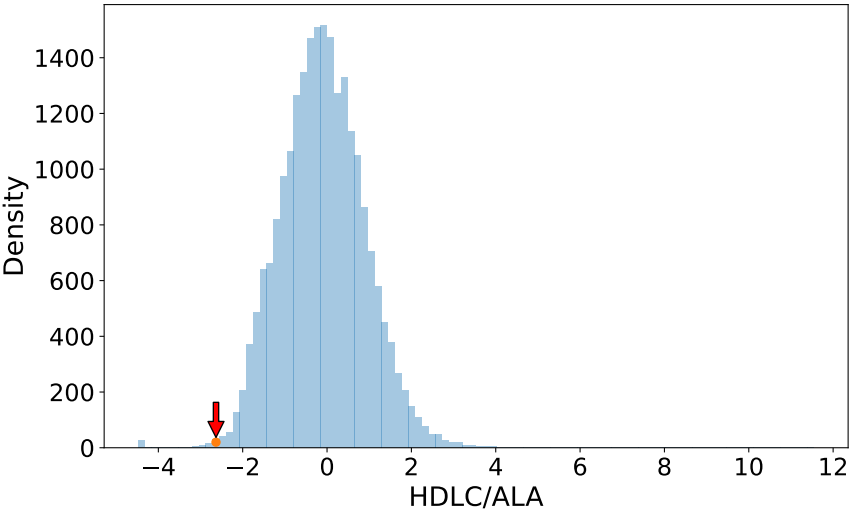}
			\put(-1,60){\large\textbf{b}}
	\end{overpic}}
	
	\subfigure{
		\label{ALA_HDLC_human_pathway}
		\begin{overpic}[width=0.65\textwidth]{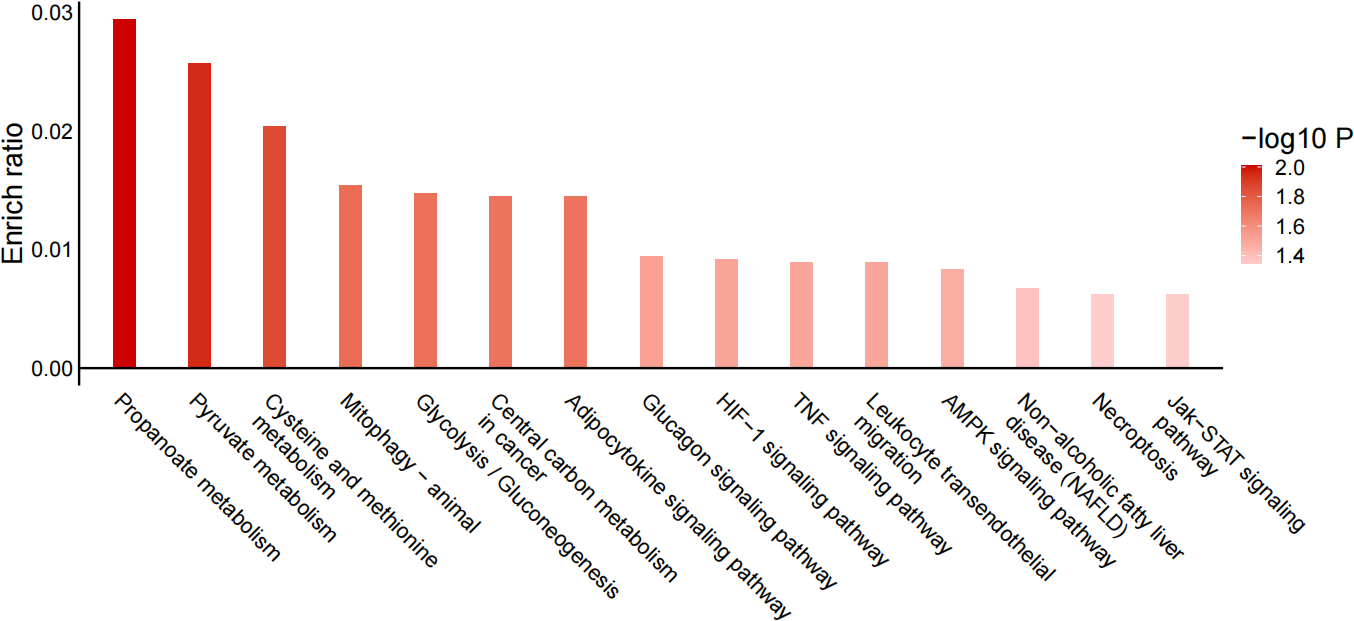}
			\put(-2,48){\large\textbf{c}}
	\end{overpic}}
	
	\subfigure{
		\label{Alb_ASA_Ap4m1}
		\begin{overpic}[width=0.25\textwidth]{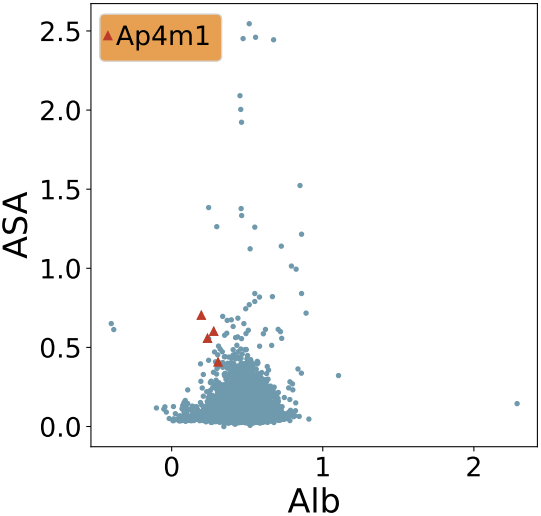}
			\put(-2,95){\large\textbf{d}}
	\end{overpic}}
	\subfigure{
		\label{Alb_ASA_ratio}
		\begin{overpic}[width=0.4\textwidth]{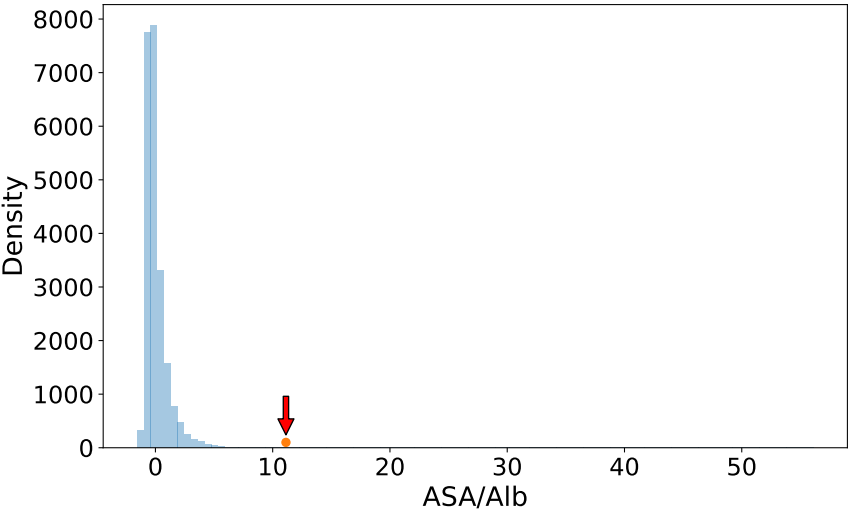}
			\put(-1,60){\large\textbf{e}}
	\end{overpic}}
	
	\subfigure{
		\label{Alb_ASA_human_pathway}
		\begin{overpic}[width=0.65\textwidth]{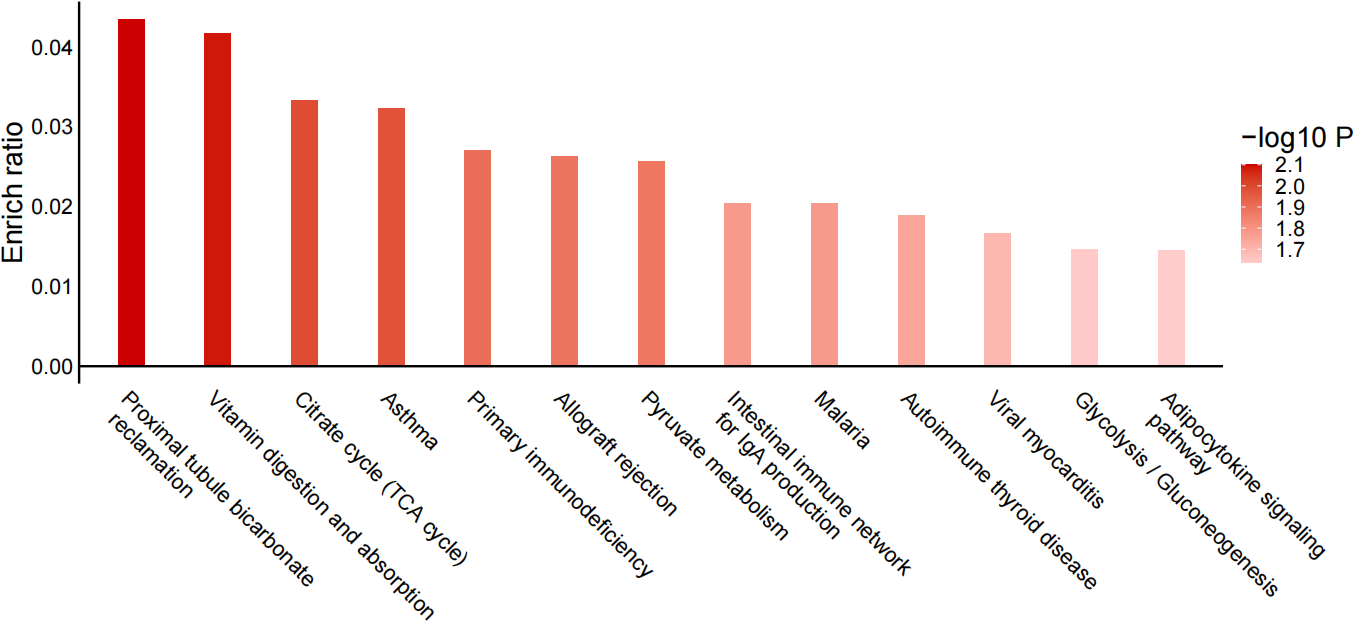}
			\put(-2,48){\large\textbf{f}}
	\end{overpic}}
	\caption{{\bf Validation of the top three abnormal parameter pairs that occur in both males and females and have the highest overall frequency.} {\bf a}, Visualization of outliers corresponding to gene \textit{Lepr} based on abnormal parameter pairs in the male dataset. {\bf b}, The distribution of the standardized ratio of ALA to HDLC in mouse metabolism-related dataset, the average ratio of \textit{Lepr} knockout mice is significantly smaller. {\bf c}, Histogram of KEGG pathway enrichment for genes corresponding to (ALA, HDLC) in humans. {\bf d}, Visualization of outliers corresponding to gene \textit{Ap4m1} based on abnormal parameter pairs in the female dataset. {\bf e}, The distribution of the standardized ratio of ASA to Alb in mouse metabolism-related dataset, the average ratio of \textit{Ap4m1} knockout mice is significantly larger. {\bf f}, Histogram of KEGG pathway enrichment for genes corresponding to (Alb, ASA) in humans.}
	\label{fig6}
\end{figure}

We then investigated the biological significance of the three most frequently perturbed parameter pairs. Previous studies have shown that the ALA/HDLC ratio is associated with the risk of non-alcoholic fatty liver disease (NAFLD) and diabetes\cite{Cao2023TheNC,He2023AlanineAT,Qiu2023TheNP}. ODBAE identified \textit{Lepr} as one of the genes causing abnormalities in this pair (Fig.\ref{fig6}a). When the ratio of ALA to HDLC was analyzed across all mice, the outliers corresponding to the \textit{Lepr} knockout had a significantly lower ratio compared to the general population (Fig.\ref{fig6}b), consistent with previous findings linking \textit{Lepr} mutations to type 2 diabetes and NAFLD\cite{Yang2016VariationsIT,Li2016GenePA}. KEGG pathway analysis of genes associated with (ALA, HDLC) revealed enrichment in pathways related to NAFLD and metabolic processes, including glycolysis (Fig.\ref{fig6}c). For the (Alb, ASA) pair, studies suggest that the ASA/Alb ratio correlates with tumor progression and prognosis in hepatocellular carcinoma (HCC)\cite{Wu2023PrognosticNF,Peng2022PreoperativeAA}. ODBAE identified \textit{Ap4m1} as a gene causing abnormalities in this pair (Fig.\ref{fig6}d). The average ratio of ASA to Alb in \textit{Ap4m1} knockout mice was significantly larger than in other mice (Fig.\ref{fig6}e), consistent with the role of \textit{Ap4m1} in HCC\cite{Peng2023AP4M1AA}. KEGG pathway enrichment further linked (Alb, ASA) disruptions to pathways involved in pyruvate metabolism, glycolysis, and proximal tubule bicarbonate reclamation (Fig.\ref{fig6}f). Finally, for genes corresponding to parameter pair (ALA, ASA), the pathway enrichment analysis revealed significant enrichment in Glycosaminoglycan biosynthesis, Starch and sucrose metabolism and Type I diabetes mellitus (Supplementary Fig.9a), consistent with the previously reported association of parameters ASA to ALA ratio with prediabetes\cite{Cao2022NonlinearRB}. In addition to these well-characterized parameter pairs, ODBAE also identified novel correlations worth exploring further. For instance, \textit{Taf8} was highlighted as a gene causing abnormalities in the parameter pair (HDLC, TP), though little is known about the specific relationship between these two parameters (Supplementary Fig.9b). Supplementary Fig.9c shows that there is a linear relationship between these two parameters, and the knockout of the gene \textit{Taf8} disrupts this relationship. ODBAE’s ability to uncover such relationships without prior biological inference makes it a powerful tool for discovering previously unknown metabolic interactions.

Overall, ODBAE systematically identifies and explains complex phenotypes by revealing intrinsic relationships between multiple metabolic parameters, including both linear and non-linear associations. These insights can serve as new indicators for biomedical research and provide a more holistic view of metabolic dysfunction in knockout mice. In addition, the ODBAE framework can be adapted to other datasets to uncover complex phenotypes in different biological systems.

\section*{Discussion}
In this study, we present ODBAE, a machine learning approach designed to detect and explain complex phenotypes by identifying outliers in high-dimensional biological datasets. The key advantage of ODBAE is its ability to detect two types of outliers: IP, which deviate significantly from expected relationships between variables, and HLP, which are far from the data center. Unlike traditional outlier detection methods that often focus on single abnormal indicators, ODBAE identifies multi-dimensional joint abnormalities, providing a more holistic view of phenotypic disruptions. This approach is particularly powerful when applied to data from knockout mouse models, where phenotypic changes are often subtle or involve complex interactions between physiological parameters.

One of the most significant contributions of ODBAE is its potential to identify unexpected phenotypes that may not be apparent when analyzing individual physiological indicators. Traditional methods of phenotype analysis often focus on abnormalities in a single trait, potentially overlooking the coordinated disruptions across multiple parameters that signal underlying biological imbalances. As our understanding of disease and homeostasis evolves, it is increasingly clear that pathological processes often involve system-wide disturbances rather than isolated dysfunctions. ODBAE leverages these insights by integrating correlations between physiological indicators, even when individual indicators remain within normal ranges. This capability allows researchers and clinicians to better understand the interplay of homeostatic mechanisms and how they are disrupted in disease states. For example, when applied to metabolism-related datasets from IMPC, ODBAE successfully identified novel metabolic genes, including those associated with complex, multi-parameter phenotypes. These phenotypes would have been difficult to detect using conventional approaches that focus on single-parameter abnormalities. ODBAE also uncovered new parameter pairs that tend to exhibit abnormalities together, providing insights into the intrinsic relationships between metabolic pathways. This type of discovery is essential for advancing our understanding of gene function and the pathophysiology of metabolic diseases.

Importantly, ODBAE offers not just detection, but also anomaly explanation through the use of kernel-SHAP. This feature allows researchers to pinpoint the exact physiological indicators driving the detected abnormalities, facilitating deeper insights into the biological processes involved. By identifying the most significant reconstruction errors and the features contributing to them, ODBAE enables a clearer interpretation of complex phenotypic data. This is particularly relevant in clinical settings, where understanding the root cause of a phenotype can guide diagnosis and treatment decisions. Although ODBAE demonstrates considerable potential, there are areas that warrant further exploration. For example, its current framework assumes a Gaussian distribution in the training data, which may limit its effectiveness in datasets that do not follow this distribution. Future work should focus on optimizing the model to handle non-Gaussian data and improving the accuracy of anomaly explanations, particularly in cases involving highly complex or subtle phenotypes.

Despite these challenges, ODBAE represents a powerful tool for the analysis of high-dimensional biomedical datasets. Its ability to detect multi-dimensional outliers, explain anomalies, and integrate correlated physiological indicators offers a new way to approach phenotypic screening, particularly in gene knockout models. We believe that ODBAE will facilitate the discovery of previously unrecognized phenotypes, enhancing our understanding of homeostasis and disease processes. By moving beyond the limitations of single-indicator analysis, ODBAE opens up new possibilities for both basic research and clinical diagnostics, allowing for a more comprehensive understanding of complex biological systems.

\subsection*{Limitation}
The current implementation of ODBAE relies on the assumption that the data follows a Gaussian distribution. However, biological data, particularly in the context of phenotypic screening, often deviates from this assumption. Non-Gaussian distributions, skewed data, and noise may impact the accuracy of outlier detection and reconstruction. Future improvements should focus on adapting the model to better handle non-Gaussian datasets, which would enhance its robustness across diverse biological and experimental contexts.

Although ODBAE could potentially be extended to other data types, such as time-series or imaging data, the current manuscript focuses exclusively on tabular datasets. The model's performance in analyzing other types of data, such as longitudinal time-series or medical imaging, has not yet been empirically tested. Further exploration and validation in these areas are warranted to assess ODBAE's broader applicability.

Additionally, ODBAE’s use in identifying phenotypes has primarily focused on metabolic parameters. Its performance in detecting phenotypes related to other physiological systems, such as neurological, cardiovascular, or developmental traits, remains unexplored. Broader testing across a range of physiological categories would demonstrate the model’s versatility and help identify any system-specific limitations.

\section*{Methods}
\subsection*{Dataset preprocessing}
We used developmental and metabolism-related datasets from IMPC to discover complex phenotypes and identify new genes. The developmental dataset we used was integrated from $8$ phenotyping centers (BCM, HMGU, ICS, JAX, MRC Harwell, RBRC, TCP and UC Davis), while the metabolism-related dataset was from $11$ phenotyping centers (BCM, CCP-IMG, HMGU, ICS, KMPC, MARC, MRC Harwell, RBRC, TCP, UC Davis and WTSI). To eliminate experiment-specific variations, we standardized each phenotyping center for each dataset to have unit variance and zero mean. Then, Min-Max normalization was applied to normalize each physiological parameter within each dataset, mitigating dimensional effects across different physiological parameters. SNPs within a $\pm$ 1 kb region of orthologues for the 41 newly identified metabolic genes were downloaded from GWAS Central on September 23, 2023 and filtered based on $-\log(p) \geq 2$.

\subsection*{Outlier detection of ODBAE}
The autoencoder model in ODBAE is formed of an encoder $f$ and a decoder $g$. If we denote the input data point and the output data point as variable $X$ and $\hat{X}$,,respectively, where $X=\big(X_1,X_2,\ldots,X_m\big)^{\top}$ and $\hat{X}=\big(\hat{X}_1,\hat{X}_2,\ldots,\hat{X}_m\big)^{\top}$, and the number of sample points is $n$, then we have $\hat{X}= g \circ f(X)$. Besides, $\mathcal{X}$ represents the whole input dataset and $X \in \mathcal{X}$. $x_i$ represents the $i^{th}$ sample point of the input dataset and its corresponding reconstruction result is $\hat{x}_i$, $x_i=\big(x_{i1},x_{i2},\ldots,x_{im}\big)^{\top}$ and $\hat{x}_i=\big(\hat{x}_{i1},\hat{x}_{i2},\ldots,\hat{x}_{im}\big)^{\top}$. Ideally, we would like to train the autoencoder with the training dataset to minimize the training loss function, the generally used training loss function is MSE which can be represented as
\begin{equation*}
	\label{L_mse}
	\begin{split}
		L_{MSE}(\omega,b)&=\frac{1}{n}\sum_{i=1}^{n}(x_i-\hat{x}_i)^{\top}(x_i-\hat{x}_i)=\frac{1}{n}\sum_{i=1}^{n}\sum_{j=1}^{m}(x_{ij}-\hat{x}_{ij})^2.
	\end{split}
\end{equation*}
Where $\omega$ represents the weight between the input layer and the output layer and $b$ is the bias value. The purpose of training process is to guarantee that the intrinsic information of the training dataset can be learned and most of the normal sample points can be well reconstructed.

\subsection*{Definition of HLP and IP}
Since the reconstruction of any dataset by the autoencoder can be regarded as a complex regression process, we can divide the outliers in each dataset into HLP and IP \cite{Chatterjee1986InfluentialOH}. For a certain dataset, we divide the dimension index into two disjoint sets $V_1$ and $V_2$. Without loss of generality, we assume that $V_1=\{1,2,\ldots,p\}$ and $V_2=\{p+1,p+2,\ldots,m\}$. For $i,j \in V_1,i \neq j$, there is no correlation between variables $X_i$ and $X_j$, while for each $k \in V_2$, there is a map $\phi_k$ such that variable $X_k=\phi_k\big(X_1,X_2,\ldots,X_p\big)$. $X_i$ $(i \in V_1)$ is called factor and $X_k$ $(k \in V_2)$ is called response variable. Then we define matrix $Q$,
\begin{equation*}
	Q=\begin{bmatrix}
		1 & x_{11} & \dots & x_{1p}\\
		1 & x_{21} & \dots & x_{2p}\\
		\vdots & \vdots & \ddots & \vdots\\
		1 & x_{n1}& \dots & x_{np}
	\end{bmatrix}.
\end{equation*} 
Besides, if we let $Q_i=\big(1,x_{i1},\ldots,x_{ip}\big)$, the leverage value of the $i^{ih}$ sample point is 
\begin{equation*}
	h_i=Q_i\big(Q^{\top}Q\big)^{-1}Q_i^{\top},\quad i=1,2,\ldots,n.
\end{equation*} 
If we represent the mean of variable $X_k$ $(k \in V_1)$ as $\mu_k$, and define the mean vector $\tilde{\mu}=\big(\mu_1,\mu_2,\ldots,\mu_p\big)$, we have
\begin{equation}
	\label{equation:xx^top}
	\begin{split}
		Q^{\top}Q &=\begin{bmatrix}
			1 & 1 & \dots & 1\\
			x_{11} & x_{21} & \dots & x_{n1}\\
			\vdots & \vdots & \ddots & \vdots\\
			x_{1p} & x_{2p}& \dots & x_{np}
		\end{bmatrix}\begin{bmatrix}
			1 & x_{11} & \dots & x_{1p}\\
			1 & x_{21} & \dots & x_{2p}\\
			\vdots & \vdots & \ddots & \vdots\\
			1 & x_{n1}& \dots & x_{np}
		\end{bmatrix}=n\begin{bmatrix}
			1 & \tilde{\mu}\\
			\tilde{\mu}^{\top} & C \\
		\end{bmatrix},
	\end{split}
\end{equation}
where $C_{jk}=\frac{1}{n}\sum_{i=1}^nx_{ij}x_{ik}$. Thus, we can obtain the inverse of $Q^{\top}Q$
\begin{equation}
	\label{equation:xx^-1}
	\begin{split}
		\big(Q^{\top}Q\big)^{-1} &=\frac{1}{n}\begin{bmatrix}
			1+\tilde{\mu}\big(C-\tilde{\mu}^{\top}\tilde{\mu}\big)^{-1}\tilde{\mu}^{\top} & -\tilde{\mu}\big(C-\tilde{\mu}^{\top}\tilde{\mu}\big)^{-1}\\
			-\big(C-\tilde{\mu}^{\top}\tilde{\mu}\big)^{-1}\tilde{\mu}^{\top} & \big(C-\tilde{\mu}^{\top}\tilde{\mu}\big)^{-1} \\
		\end{bmatrix}.
	\end{split}
\end{equation}
Therefore, if we let $\tilde{x}_i=\big(x_{i1},\ldots,x_{ip}\big)$, the leverage value of the $i^{th}$ sample point can also be formulated as
\begin{equation}
	\label{equation:hi}
	\begin{split}
		h_i &=\frac{1}{n}Q_i\begin{bmatrix}
			1+\tilde{\mu}\big(C-\tilde{\mu}^{\top}\tilde{\mu}\big)^{-1}\tilde{\mu}^{\top} & -\tilde{\mu}\big(C-\tilde{\mu}^{\top}\tilde{\mu}\big)^{-1}\\
			-\big(C-\tilde{\mu}^{\top}\tilde{\mu}\big)^{-1}\tilde{\mu}^{\top} & \big(C-\tilde{\mu}^{\top}\tilde{\mu}\big)^{-1} \\
		\end{bmatrix}Q_i^{\top}\\
		&=\frac{1}{n}\Big(1+\big(\tilde{\mu}-\tilde{x}_i\big)\big(C-\tilde{\mu}^{\top}\tilde{\mu}\big)^{-1}\tilde{\mu}^{\top},\big(\tilde{x}_i-\tilde{\mu}\big)\big(C-\tilde{\mu}^{\top}\tilde{\mu}\big)^{-1}\Big)Q_i^{\top}\\
		&=\frac{1}{n}\Big(1+\big(\tilde{\mu}-\tilde{x}_i\big)\big(C-\tilde{\mu}^{\top}\tilde{\mu}\big)^{-1}\tilde{\mu}^{\top}+\big(\tilde{x}_i-\tilde{\mu}\big)\big(C-\tilde{\mu}^{\top}\tilde{\mu}\big)^{-1}\tilde{x}_i^{\top}\Big)\\
		&=\frac{1}{n}+\frac{1}{n}\big(\tilde{x}_i-\tilde{\mu}\big)\big(C-\tilde{\mu}^{\top}\tilde{\mu}\big)^{-1}\big(\tilde{x}_i-\tilde{\mu}\big)^{\top}.
	\end{split}
\end{equation}
If we define $A=C-\tilde{\mu}^{\top}\tilde{\mu}$, then
\begin{equation}
	\label{equation:A_jk}
	\begin{split}
		A_{jk} &= \frac{1}{n}\sum_{i=1}^{n}x_{ij}x_{ik}-\mu_j\mu_k\\
		&= \frac{1}{n}\sum_{i=1}^{n}\big(x_{ij}-\mu_j\big)\big(x_{ik}-\mu_k\big) \\
		&= \frac{n-1}{n}COV(X_j,X_k),\quad j,k \in V_1. \\
	\end{split}
\end{equation}
Furthermore, we have $A=\frac{n-1}{n}\Sigma_{f}$ and $A^{-1}=\frac{n}{n-1}\Sigma_{f}^{-1}$, where $\Sigma_{f}$ is the covariance matrix of the factor space. Then, based on equation (\ref{equation:hi}), we get
\begin{equation}
	\label{equation:hi2}
	\begin{split}
		h_i &= \frac{1}{n}+\frac{1}{n-1}\big(\tilde{x}_i-\tilde{\mu}\big)\Sigma_f^{-1}\big(\tilde{x}_i-\tilde{\mu}\big)^{\top}.
	\end{split}
\end{equation}
Obviously, for each sample point, its leverage value is proportional to its Mahalanobis distance in the factor space. In addition, the Cook distance of the $i^{th}$ sample point based on the $k^{th}$ $(k \in V_2)$ dimension is
\begin{equation}
	\begin{split}
		C_{ik}=\frac{(n-p)\big(x_{ik}-\hat{x}_{ik}\big)^2h_i}{p\sum_{i=1}^{n}\big(x_{ik}-\hat{x}_{ik}\big)^2\big(1-h_i\big)^2},\quad i=1,2,\ldots,n.
	\end{split}
\end{equation}

For each dataset, HLP are sample points with high leverage value and IP refer to sample points with high Cook distance. The leverage value of the $i^{th}$ sample point is also proportional to the Mahalanobis distance of the factor space. Therefore, HLP can also be identified by the Mahalanobis distance in the factor space. It is noteworthy that for certain datasets, like those conforming to a multivariate Gaussian distribution, the set $V_2$ may be empty. In such instances, all outliers in the dataset are identified as HLP.

\subsection*{Training loss function of ODBAE}
To explain why MSE-trained autoencoders are insufficient in outlier detection, we presented rigorous theoretical analysis and further analyzed how to solve these problems. The analysis process is based on the following assumption.
\newtheorem{assumption}{Assumption}
\begin{assumption}
	We assume that the input variable $X$ follows m-dimensional Gaussian distribution.
	\label{assumption1}
\end{assumption}

As we know, autoencoders will mainly focus on recovering the principal components of a dataset eventually \cite{journals/corr/abs-2007-06731,oftadeh2020eliminating}, it is necessary to use the unit orthogonal eigenvectors as the basis vectors to further study the influence of the loss function on the outlier detection results of the autoencoder. Under Assumption \ref{assumption1}, we represent the mean vector of the variable $X$ as $\mu=\big(\mu_1,\mu_2,\ldots,\mu_m\big)^{\top}$. If the eigenvalues of the covariance matrix $\Sigma_x$ are $\lambda_1$, $\lambda_2$, $\ldots$, $\lambda_m$, the corresponding unit orthogonal eigenvectors (i.e., principal directions) are $\eta_1$, $\eta_2$, $\ldots$, $\eta_m$, and the new coordinate of the variable $X$ is $Y=\big(Y_1,Y_2,\ldots,Y_m\big)^{\top}$  when the unit orthogonal eigenvectors are the basis vectors, then we can denote an orthogonal matrix $P$, where the $i^{th}$ column of $P$ is unit eigenvector $\eta_i$. It's obvious that $Y=P^{-1}X=P^{\top}X$ and $Y_k=\eta_k^{\top}X$, $k=1,2,\ldots,m$. Thus, in the new coordinate system, the variable of input data $X$ is converted to variable $Y$. Let $\nu=\big(\nu_1,\nu_2,\ldots,\nu_m\big)^{\top}$ and $\Sigma_Y$ represent the mean vector and covariance matrix of variable $Y$, respectively, we have $\nu$ $=$ $P^{\top}\mu$ and $\nu_k$ $=$ $E(Y_k)$ $=$ $\eta_k^{\top}\mu$, $k$ $=$ $1,2,\ldots,m$. Furthermore, we can gain the variance of the $k^{th}$ element of variable $Y$ is
\begin{equation}
	\label{equation:D(Y)}
	\begin{split}
		D(Y_k) &= COV\big(Y_k,Y_k\big)\\
		&= E\Big(\big(Y_k-\nu_k\big)\big(Y_k-\nu_k\big)\Big) \\
		&= E\Big(\eta_k^{\top}\big(X-E(X)\big)\eta_k^{\top}\big(X-E(X)\big)\Big) \\
		&= \eta_k^{\top}E\Big(\big(X-E(X)\big)\big(X-E(X)\big)^{\top}\Big)\eta_k \\
		&= \eta_k^{\top}\Sigma_x\eta_k \\
		&= \eta_k^{\top}\lambda_k\eta_k \\
		&= \lambda_k,
	\end{split}
\end{equation}
and the covariance of $Y_i$ and $Y_j$ ($i$ $\neq$ $j$) is
\begin{equation}
	\label{equation:COV(Yi,Yj)}
	\begin{split}
		COV(Y_i,Y_j) &= E\Big(\big(Y_i-E(Y_i)\big)\big(Y_j-E(Y_j)\big)\Big)\\
		&= E\Big(\eta_i^{\top}\big(X-E(X)\big)\eta_j^{\top}\big(X-E(X)\big)\Big) \\
		&= \eta_i^{\top}E\Big(\big(X-E(X)\big)\big(X-E(X)\big)^{\top}\Big)\eta_j \\
		&= \eta_i^{\top}\Sigma_x\eta_j \\
		&= \eta_i^{\top}\lambda_j\eta_j \\
		&= 0.
	\end{split}
\end{equation}
Thus, the covariance matrix of variable $Y$ is $\Sigma_Y={\rm diag}(\lambda_1,\lambda_2,\ldots,\lambda_m)$.
\begin{thm}
	The covariance matrix of variable $Y$ can be derived as $$\Sigma_Y={\rm diag}(\lambda_1,\lambda_2,\ldots,\lambda_m).$$
	\label{Result1}
\end{thm}
If we denote $\nu=(\nu_1,\nu_2,\ldots,\nu_m)^{\top}$ as the mean vector of variable $Y$, since $X$ follows m-dimensional Gaussian distribution, we can conclude that $Y$ also follows m-dimensional Gaussian distribution, i.e., $Y$ $\sim$ $\mathcal{N} (\nu,\Sigma_Y)$. What's more, it's easy to get that $Y_k$ $\sim$ $\mathcal{N} (\nu_k,\lambda_k)$, $k=1,2,\ldots,m$. Then, we denote $\hat{Y}$ as the output variable of $Y$ and its mean vector and eigenvalues of the covariance matrix are $\hat{\nu}=\big(\hat{\nu}_1,\hat{\nu}_2,\ldots,\hat{\nu}_m\big)^{\top}$ and $\hat{\lambda}_k$, $k=1,2,\ldots,m$, respectively. As is mentioned above, autoencoders will mainly recover the principal components of any datasets, it's reasonable for us to make the following assumption.

\begin{assumption}
	\label{assumption2}
	For each element $\hat{Y}_k$ of the output variable $\hat{Y}$, its data distribution is the same as the corresponding input element $Y_k$. Since $Y_k$ $\sim$ $\mathcal{N} (\nu_k,\lambda_k)$, then $\hat{Y}_k$ $\sim$ $\mathcal{N} (\hat{\nu}_k,\hat{\lambda}_k)$, $k=1,2,\ldots,m$. Besides, whether each eigenvalue after reconstruction is close to 0 is the same as its corresponding original eigenvalue.
\end{assumption}

In the following, we will theoretically analyze the restrictions of MSE in detecting HLP. First, we can get the relationship between the difference between the reconstructed data and its mean and the difference between the original data and its mean in each principal component direction. 

As we know, the loss functhion of MSE can be formulated as
\begin{equation}
	\label{equation:MSE}
	\begin{split}
		L_{MSE}(\omega,b) &= \frac{1}{n}\sum_{i=1}^{n} (y_i-\hat{y}_i)^{\top}(y_i-\hat{y}_i)\\
		&= E\big(Y^{\top}Y-2Y^{\top}\hat{Y}+\hat{Y}^{\top}\hat{Y}\big)\\
		&= E\big(Y^{\top}Y\big)-2E\big(Y^{\top}\hat{Y}\big)+E\big(\hat{Y}^{\top}\hat{Y}\big) \\
		&= \sum_{i=1}^m E\big(Y_i^2\big)-2\sum_{i=1}^m E\big(Y_i\hat{Y}_i\big)+\sum_{i=1}^m E\big(\hat{Y}_i^2\big), \\
	\end{split}
\end{equation}
where $y_i$ and $\hat{y}_i$ represent the $i^{th}$ sample point of the input and output dataset in the new coordinate system respectively.

First, if the autoencoder can properly reconstruct the input dataset, we will analyse the mean vector of the output variable $\hat{Y}$. Based on Assumption \ref{assumption2} and equation (\ref{equation:MSE}), we have
\begin{equation}
	\label{equation:MSE case 2}
	\begin{split}
		L_{MSE}(\omega,b) &= \sum_{i=1}^m \lambda_i+\sum_{i=1}^m \hat{\lambda}_i+\sum_{i=1}^m \nu_i^2+\sum_{i=1}^m \hat{\nu}_i^2-2\sum_{i=1}^m \nu_i\hat{\nu}_i-2\sum_{i=1} ^m COV\big(Y_i,\hat{Y}_i\big)\\
		&\geq \sum_{i=1}^m \big(\nu_i-\hat{\nu}_i\big)^2+\sum_{i=1}^m \Bigg(\sqrt{\lambda_i}-\sqrt{\hat{\lambda}_i}\Bigg)^2. \\
	\end{split}
\end{equation}
In order to reach the minima of $L_{MSE}(\omega,b)$, the mean vector of output variable $\hat{\nu}$ must satisfy $\hat{\nu}_k=\nu_k$, $k$ $=$ $1,2,\ldots,m$. That is to say, the mean vector of output data is the same as input data.

Next, we will specifically analyze the reconstruction error of each data point. According to Assumptions \ref{assumption1} and \ref{assumption2}, the input and output variable satisfy $Y_k$ $\sim$ $\mathcal{N} (\nu_k,\lambda_k)$ and $\hat{Y}_k$ $\sim$ $\mathcal{N} (\nu_k,\hat{\lambda}_k)$, $k=1,2,\ldots,m$. Denote $R(Y)=\big(R_1(Y),R_2(Y),\ldots,R_m(Y)\big)^{\top}$ \big($\hat{R}(Y)=\big(\hat{R}_1(Y),\hat{R}_2(Y),\ldots,\hat{R}_m(Y)\big)^{\top}$\big) as the difference between variable $Y$ \big($\hat{Y}$\big) and its mean vector $\nu$, i.e., $R(Y)=Y-\nu$ and $\hat{R}(Y)=\hat{Y}-\nu$. It's easy to see that $R_k(Y)$ $\sim$ $\mathcal{N} (0,\lambda_k)$ and $\hat{R}_k(Y)$ $\sim$ $\mathcal{N} (0,\hat{\lambda}_k)$. Then the following equation can be obtained.
\begin{equation}
	\label{equation:MSE case 3}
	\begin{split}
		L_{MSE}(\omega,b) &=E\Big(\big(Y-\hat{Y}\big)^{\top}\big(Y-\hat{Y}\big)\Big)\\
		&=E\Big(\big(R(Y)-\hat{R}(Y)\big)^{\top}\big(R(Y)-\hat{R}(Y)\big)\Big) \\
		&=E\Big(R^{\top}(Y)R(Y)\Big)-2E\Big(R^{\top}(Y)\hat{R}(Y)\Big)+E\Big(\hat{R}^{\top}(Y)\hat{R}(Y)\Big) \\
		&=\sum_{i=1}^mE\Big(R_i^2(Y)\Big)+\sum_{i=1}^mE\Big(\hat{R}_i^2(Y)\Big)-2\sum_{i=1}^mE\Big(R_i(Y)\Big)E\Big(\hat{R}_i(Y)\Big)\\
		&\hskip 0.5cm -2\sum_{i=1}^mCOV\Big(R_i(Y),\hat{R}_i(Y)\Big) \\
		&=\sum_{i=1}^m\lambda_i+\sum_{i=1}^m \hat{\lambda}_i-2\sum_{i=1}^m \rho_i\sqrt{\lambda_i}\sqrt{\hat{\lambda}_i},
	\end{split}
\end{equation}
where $\rho_k$ represents the correlation coefficient of $R_k(Y)$ and $\hat{R}_k(Y)$, and $-1 \leq \rho_k \leq 1$, $k=1,2,\ldots,m$. In equation (\ref{equation:MSE case 3}), only when $\rho_k=1$, can $L_{MSE}(\omega,b)$ reach its minima. In this case, there is a positive linear correlation between $R_k(Y)$ and $\hat{R}_k(Y)$, specifically, there exist constants $a_k$ and $c_k$ such that $\hat{R}_k(Y)=a_kR_k(Y)+c_k$, then $E\big(\hat{R}_k(Y)\big)=a_kE\big(R_k(Y)\big)+c_k$ and $D\big(\hat{R}_k(Y)\big)=a_k^2D\big(R_k(Y)\big)$. Since $E\big(\hat{R}_k(Y)\big)=E\big(R_k(Y)\big)=0$, $D\big(R_k(Y)\big)=\lambda_k$ and $D\big(\hat{R}_k(Y)\big)=\hat{\lambda}_k$, we can obtain $c_k=0$ and $a_k=\frac{\sqrt{\hat{\lambda}_k}}{\sqrt{\lambda_k}}$. Therefore, we can obtain the following proposition.

\begin{thm}
	\label{Result2}
	Under Assumptions \ref{assumption1} and \ref{assumption2}, the difference between the reconstructed data and its mean is proportional to the difference between the original data and its mean in each principal component direction. The specific relationship is 
	\begin{equation}
		\label{equation:R_k(Y)}
		\begin{split}
			\hat{R}_k(Y)=\frac{\sqrt{\hat{\lambda}_k}}{\sqrt{\lambda_k}}R_k(Y).
		\end{split}
	\end{equation}
\end{thm}
Where $R_k(Y)=Y_k-\nu_k$ and $\hat{R}_k(Y)=\hat{Y}_k-\hat{\nu}_k$. Based on equation (\ref{equation:R_k(Y)}), we can further determine the specific reconstruction error of each input data point.

\begin{thm}
	\label{Result3}
	The reconstruction error of each input data point measured by MSE can be formulated as
	\begin{equation}
		\label{equation:reconstruct error}
		\begin{split}
			W(Y) &=\big(Y-\hat{Y}\big)^{\top}\big(Y-\hat{Y}\big)\\
			&=\sum_{i=1}^m\Bigg(\sqrt{\lambda_i}-\sqrt{\hat{\lambda}_i}\Bigg)^2\frac{R_i^2(Y)}{\lambda_i}.\\
		\end{split}
	\end{equation}
\end{thm}

Ideally, we hope each principal component contributes equally to the reconstruction error of any input data point. Recall the definition of $R(Y)$, it's easy to know that $\frac{R_k(Y)}{\lambda_k}$ $\sim$ $\mathcal{N} (0,1)$, so equation (\ref{equation:reconstruct error}) indicates that the reconstruction degree of each principal component by the autoencoder will affect the proportion of this principal component in the reconstruction error. This brings a lot of uncertainty for HLP detection.

\begin{thm}
	\label{Result4}
	HLP are difficult to identify as outliers if the input dataset is completely reconstructed.
\end{thm}

When the input dataset is well reconstructed, in equation (\ref{equation:reconstruct error}), it means that $\sqrt{\lambda_k}-\sqrt{\hat{\lambda}_k}=0$, $k$ $=$ $1,2,\ldots,m$. Then, the reconstruction error of each data point in training dataset is $0$. As the result, each data point in the training dataset is considered normal. However, HLP exist in any dataset follows m-dimensional Gaussian distribution. Generally, the larger the Mahalanobis distance is, the more abnormal the data point is. Therefore, although completely reconstructing the training dataset is beneficial for IP detection, it ignores the detection of HLP.

\begin{thm}
	\label{Result5}
	HLP whose values in each dimension are within the normal range can not be identified as outliers. Besides, most HLP will be detected in the direction corresponding to the worst-recovered principal component, but in the direction of the well-recovered principal components, the anomalies are often ignored.
\end{thm}

From equation (\ref{equation:reconstruct error}), we know $\sqrt{\lambda_k}-\sqrt{\hat{\lambda}_k}$ is equivalent to the weight of the reconstruction error of the $k^{th}$ principal component direction of each data point to the total reconstruction error. Therefore, if the data reconstruction in the $i^{th}$ principal component direction is better than that in the $j^{th}$ principal component direction, then $\sqrt{\lambda_i}-\sqrt{\hat{\lambda}_i}<\sqrt{\lambda_j}-\sqrt{\hat{\lambda}_j}$, $i \neq j$. As the result,  the detected outliers are more distributed in the $j^{th}$ principal component direction.

\begin{thm}
	\label{Result6}
	If we ensure the differences between the eigenvalues of the covariance matrix of the original dataset and their corresponding reconstructed results in the direction of each principal component are equal, the value of the reconstruction error for each data point will be proportional to its Mahalanobis distance.
\end{thm}
If we suppress complete reconstruction of the autoencoder, then $\sqrt{\lambda_k}-\sqrt{\hat{\lambda}_k}>0$, $k=1,2,\ldots,m$. Based on equation (\ref{equation:R_k(Y)}), we can obtain
\begin{equation}
	\label{equation:reconstruct_error_Y_k}
	\begin{split}
		(Y_k-\hat{Y}_k)^2=\frac{\Big(\sqrt{\lambda_k}-\sqrt{\hat{\lambda}_k}\Big)^2}{\lambda_k}(Y_k-\nu_k)^2.
	\end{split}
\end{equation}
It means that for each principal component direction of the input data, there is a reconstruction error. Specifically, the further away the values are from the mean, the larger the reconstruction error. Therefore, in each principal component direction, the outliers detected by the reconstruction error are the same as the outliers defined by the Gaussian distributed data. Besides, if $\sqrt{\lambda_k}-\sqrt{\hat{\lambda}_k}=\beta>0$, $k=1,2,\ldots,m$, according to equation (\ref{equation:reconstruct error}), the reconstruction error of each input data point is
\begin{equation}
	\label{equation:reconstruct error control eig}
	\begin{split}
		W(Y) &=\big(Y-\hat{Y}\big)^{\top}\big(Y-\hat{Y}\big)=\beta^2\sum_{i=1}^m\frac{R_i^2(Y)}{\lambda_i}.\\
	\end{split}
\end{equation}
Since $\Sigma_Y={\rm diag}(\lambda_1,\lambda_2,\ldots,\lambda_m)$, it's easy to obtain $\Sigma_Y^{-1}={\rm diag}\Big(\frac{1}{\lambda_1},\frac{1}{\lambda_2},\ldots,\frac{1}{\lambda_m}\Big)$. So the Mahalanobis distance of the input variable $Y$ is
\begin{equation}
	\label{equation:Mahalanobis distance}
	\begin{split}
		M(Y) &=\big(Y-\nu\big)^{\top}\Sigma_Y^{-1}\big(Y-\nu\big)=\sum_{i=1}^m\frac{R_i^2(Y)}{\lambda_i}.
	\end{split}
\end{equation}

It indicates that the reconstruction error of each data point is proportional to its Mahalanobis distance. Therefore, if we control $\sqrt{\lambda_k}-\sqrt{\hat{\lambda}_k}=\beta>0$, $k=1,2,\ldots,m$, almost all HLP will be converted to IP and cannot be well reconstructed. As the result, the detected HLP will evenly distributed in each principal component direction.

Based on the above discussion, we will propose a new loss function that adds an appropriate penalty term based on MSE to balance the reconstruction of the autoencoder.

In fact, if the intrinsic dimension of the dataset is $l$, then there will be $l$ eigenvalues that are not close to $0$. We denote the $l$ eigenvalues that are not close to $0$ as $\lambda_k^{'}$ and their corresponding reconstruction results are $\hat{\lambda}_k^{'}$,  $k=1,2,\ldots,l$. According to Assumption \ref{assumption2}, $\hat{\lambda}_k^{'}$ is also not close to $0$. Then, we consider two losses in our training loss function. One of them is $L_{MSE}(\omega,b)$, which aims to reconstruct the input dataset well. In addition, we define the other loss $L_{EIG}(\omega,b)=\sum_{i=1}^l\big(\sqrt{\lambda_i^{'}}-\sqrt{\hat{\lambda}_i^{'}}-\beta\big)^2$ which can avoid the autoencoder from completely reconstructing the input dataset. $\beta$ $>$ $0$ is a hyperparameter that can adjust the degree of data reconstruction. The final training loss function is a combination of the two:
\begin{equation}
	\label{ODBAE_Loss}
	\begin{split}
		L(\omega,b)=\theta_1L_{MSE}(\omega,b)+\theta_2L_{EIG}(\omega,b).
	\end{split}
\end{equation}
Here, $\theta_1$, $\theta_2$ $>$ $0$ are hyperparameters that need to be predetermined. In practice, desirable results are usually obtained if we set $\theta_1=0.008$ and $\theta_2=1$.

\subsection*{Anomaly explanation of ODBAE}
For each outlier detected by ODBAE, it's necessary for us to find its abnormal parameters. 
For ODBAE, it obtained abnormal parameters based on the highest reconstruction errors and SHAP values. 
Specifically, for each outlier, we first sorted its parameters in descending order based on the reconstruction errors. Then, if the sum of the reconstruction errors of the first $n$ parameters was greater than $50$\% of the total reconstruction error for the outlier, these $n$ parameters were considered potentially anomalous. If $n=1$, we used SHAP values to identify additional parameter that have the greatest impact on this anomalous parameter. If the SHAP value of these two parameters exceeded the set threshold, they would be considered abnormal parameters; otherwise, only the parameter with the highest reconstruction error was considered an abnormal parameter. However, if $n>1$, we would use SHAP values to obtain additional parameter that have the greatest impact on the parameter with the highest reconstruction error. If the SHAP value of these two parameters exceed the set threshold, they would be considered abnormal parameters; otherwise, the top $2$ parameters with the highest reconstruction error were considered anomalous parameters. In our work, the threshold for SHAP values was set to the mean of the SHAP values during the process of anomaly explanation for all outliers.

\subsection*{Identification of outliers using z-score method}
In developmental and metabolism-related datasets, we first identified mice with anomalies using the z-score method. For example, there were $14$ metabolic parameters in metabolism-related datasets. Then, the z-score value of each mouse for each metabolic parameter was calculated as $Z_{ij}=\frac{X_{ij}-\mu_i}{\sigma_i}$, where $Z_{ij}$ represented the z-score of the $j^{th}$ mouse for the $i^{th}$ metabolic parameter, $X_{ij}$ was the value of the $i^{th}$ metabolic parameter for the $j^{th}$ mouse, $\mu_i$ and $\sigma_i$ were the mean and standard deviation of the $i^{th}$ metabolic parameter, respectively. The larger the absolute value of $Z_{ij}$, the more the $i^{th}$ metabolic parameter value of the $j^{th}$ mouse deviates from the mean. Therefore, for each metabolic parameter, mice with large absolute z-score values were labeled as outliers.

\subsection*{Identification of mice with low BMI values}
We calculated the BMI values for all knockout mice, and the mean BMI of the abnormal mice after \textit{Ckb} knockout in female mice was lower than that of $97.14$\% of the mice. Therefore, there were still a small number of mice with lower BMIs from a total of $264$ single-gene knockout mouse strains, and only $13$ ($4.92$\%) single-gene knockout mouse strains had more than $50$\% of mice with BMI values lower than the mean BMI of gene \textit{Ckb} knockout mice. Among these $13$ genes, $7$ genes were identified as important genes by ODBAE, and for the remaining $6$ genes, a portion of mice from each corresponding gene knockout strain were detected as outliers.

\subsection*{Impact of the saturating nature of the non-linear activation function}
Usually, the dataset is normalized before being fed into the model, so the values of each dimension of the dataset are distributed in $(0,1)$. If we denote $F= g \circ f$, $F=(F_1,F_2,\ldots,F_m)^{\top}$, since the output variable $\hat{X}$ and the input variable $X$ satisfy that $\hat{X} = F(X)$, for each dimension of the input variable, its corresponding reconstructuion result can be fomulated as $\hat{X}_k=F_k(X_1,X_2,\cdots,X_m)$, $k=1,2,\ldots,m$. When the input dataset obeys m-dimensional Gaussian distribution, the mean reconstruction error of each dimension is
\begin{equation}
	\label{equation:X_k^_X_k}
	\begin{split}
		E\big((X_k-\hat{X}_k)^2\big) &=E\big(X_k^2\big)-2E\big(X_k\hat{X}_k\big)+E\big(\hat{X}_k^2\big)\\
		&=\mu_k^2+\sigma_k^2-2\mu_k\hat{\mu}_k-2COV\big(X_k,\hat{X}_k\big)+\hat{\mu}_k^2+\hat{\sigma}_k^2\\
		&=\big(\mu_k-\hat{\mu}_k\big)^2+\big(\sigma_k^2-2\rho_k\sigma_k\hat{\sigma}_k+\hat{\sigma}_k^2\big).
	\end{split}
\end{equation} 
Where $\mu_k$ denotes the mean of variable $X_k$ and its corresponding reconstruction result is $\hat{\mu}_k$. $\sigma_k$ represents the variance of variable $X_k$ and its corresponding reconstruction result is $\hat{\sigma}_k$. Besides, $COV\big(X_k,\hat{X}_k\big)$ is the covariance of $X_k$ and $\hat{X}_k$, $\rho_k \in [-1,1]$ is the correlation coefficient between $X_k$ and $\hat{X}_k$. To minimize the mean of the reconstruction error for the $k^{th}$ dimension, we have $\rho_k=1$, which means that there exist constants $a>0$ and $b$ such that $\hat{X}_k=aX_k+b$.

Therefore, for each dimension of the input dataset, $\hat{X}_k$ increases monotonically with $X_k$ and $\hat{X}_k\approx X_k$ holds for most data points after reconstruction. As the output layer of the autoencoder uses the sigmoid activation function, $\hat{X}_k$ is output through the sigmoid activation function. Due to the saturation of the non-linear activation functions, in each dimension, values close to $0$ or close to $1$ are more difficult to reconstruct, and $\hat{X}_k$ cannot be approximated to $X_k$ in this case.

\subsection*{Comprehensive performance evaluation}
We compared ODBAE with MSE-AE, MAE-AE and DAGMM for the detection results of IP and HLP and computed their AUC and AP scores by regarding the IP and HLP as positive.

We implemented DAGMM with minimal modification so that it adapt to our datasets. Besides, for DAGMM, we kept its number of layers and the dimensions of each layer consistent with ODBAE, MSE-AE and MAE-AE.

For each experiment, the structure of the autoencoder was as follows. The dimension of the hidden layer was the intrinsic dimension of the input dataset, and its activation function was relu. Besides, the activation function of the output layer was sigmoid. All datasets were normalized before outlier detection. During the autoencoder training process, Adam\cite{Kingma2014AdamAM} was used to optimize the loss function and the learning rate was $10^{-3}$. For our new training loss function, we always set $\theta_1=0.008$ and $\theta_2=1$.

\subsection*{Selection of hyperparameters $\theta_1$, $\theta_2$ in training loss function}
Our new training loss function is formulated as $L(\omega,b)=\theta_1L_{MSE}(\omega,b)+\theta_2L_{EIG}(\omega,b)$. It can be seen that our loss function contains two loss terms. The function of $L_{MSE}(\omega,b)$ is to drive the autoencoder to reconstruct the input dataset as well as possible, while $L_{EIG}(\omega,b)$ aims at suppressing complete reconstruction of the dataset by the autoencoder and balance the degree of reconstruction in the direction of each principal component of the input dataset. Obviously, when the value of $\theta_1:\theta_2$ is large, the autoencoder can reconstruct the input dataset perfectly and will adversely affect the detection of HLP. However, if the value of $\theta_1:\theta_2$ is very small, the autoencoder will have a poor reconstruction effect on the input dataset, which will bring disadvantages to IP detection. Therefore, it is very necessary for us to ensure that the value of $\theta_1:\theta_2$ is reasonable, so that the autoencoder can have a good detection effect for HLP and IP at the same time.

In our work, we found that when $\theta_1:\theta_2=0.008:1$, all experiments could obtain the desired results. In order to study the sensitivity of this ratio, we changed its base to see how different base affects the accuracy of outlier detection. To be specific, if the value of base was $c$, we set the values of $\theta_1$ and $\theta_2$ to $0.008c$ and $c$, respectively.

Table \ref{tab2} shows the AUC and AP scores of the Dry Bean Dataset detected by ODBAE. It's easy to see that the AUC and AP scores varied very little when the value of the base changed, which indicated that $\theta_1$ and $\theta_2$ were not sensitive to the change of the base.
\begin{table}[h]
	\centering
	\caption{Sensitivity of $\theta_1$ and $\theta_2$ with fixed ratio 0.008:1 on Dry Bean Dataset.}\label{tab2}%
	\begin{tabular}{@{}c|cccr@{}}
		\toprule
		\makebox[0.15\textwidth][c]{Base} & \makebox[0.2\textwidth][c]{AUC score} & \makebox[0.2\textwidth][c]{AP score} \\
		\midrule
		1    & 0.937972 & 0.787625 \\
		3    & 0.940508 & 0.790070 \\
		5    & 0.933077 & 0.798889 \\
		7    & 0.939188 & 0.798299 \\
		9    & 0.932391 & 0.788327 \\
		\bottomrule
	\end{tabular}
\end{table}

\subsection*{Determination of hyperparameter $\beta$}
Although hyperparameter $\beta>0$ in equation (\ref{ODBAE_Loss}) is beneficial for HLP detection, as the value of $\beta$ increases, the data reconstruction ability of the autoencoder will become worse and worse, which is similar to the model not being well fitted during regression analysis. This adversely affects the detection of IP. Therefore, in the following, it's important for us to determine how to choose the appropriate value of $\beta$.

If the intrinsic and actual dimensions of the dataset are equal, i.e., $l=m$, then most of the outliers in the dataset are HLP. In this case, considering that $\sqrt{\lambda_k}-\sqrt{\hat{\lambda}_k}>0$ and $\hat{\lambda}_k>0$, $k$ $=$ $1,2,\ldots,m$, the value of the hyperparameter $\beta$ should satisfy $0<\beta<\mathop{\min}_{1 \leq i \leq m}(\sqrt{\lambda_i})$. According to the previous analysis, as long as $\beta>0$, the detection effect of HLP can be improved. Meanwhile, in order to maintain the detection effect of IP, the value of $\beta$ must be small enough.

According to equation (\ref{equation:R_k(Y)}), if $\beta=0$, then $\hat{R}_k(Y)=R_k(Y)$, i.e., $\hat{Y}_k=Y_k$, $k$ $=$ $1,2,\ldots,m$. However, we usually add nonlinear activation functions to the autoencoder to learn nonlinear features in the input dataset. For example, in our work, we added sigmoid activation function to the output layer. Since commonly used nonlinear activation functions are saturated, when the input values are very large or very small, the corresponding output values hardly change with the change of the input values. As the result, for each dimension of the input data, when the values are very large or very small, their reconstruction results are poor. Otherwise, the values can be reconstructed well. Then, most HLP detected by the autoencoder are anomalous in one dimension, but HLP whose values in each dimension are within the normal range are ignored. It can be seen from this that when $\beta$ is small enough, it not only adversely affects the detection effect of HLP, but also rarely improves the detection effect of IP. That is to say, there exists $\xi>0$, when $\beta<\xi$, there is not much improvement in the detection effect of IP.

If we determine the value of $\xi$ and let $\beta=\xi$, on the one hand, the detection effect of the autoencoder for IP can be maintained. On the other hand, we can balance the reconstruction of the autoencoder and avoid the influence of the saturation of the activation function on the data reconstruction at the same time, which can improve the detection effect for HLP.

In our work, the structure of the autoencoder and the setting of hyperparameters are described above. For each dimension of any dataset, input values between $0.15$ and $0.85$ are hardly affected by the saturation of the nonlinear activation function. Therefore, we hope that the value of $\beta$ selected can make the output value of each dimension between $0.15$ and $0.85$. As the result, for each dimension, the interval length of the output value is $0.7$ times the interval length of the input value. Before outlier detection, the input dataset is normalized, so the input value of each dimension is between $0$ and $1$. Recall that $Y_k$ $\sim$ $\mathcal{N} (\nu_k,\lambda_k)$ and $\hat{Y}_k$ $\sim$ $\mathcal{N} (\nu_k,\hat{\lambda}_k)$, $k=1,2,\ldots,m$, the probability that $Y_k$ is distributed in $(\nu_k-3\sqrt{\lambda_k}, \nu_k+3\sqrt{\lambda_k})$ is $0.9974$, so its distribution interval length is approximately $6\sqrt{\lambda_k}$. Similarly, the distribution interval length of $\hat{Y}_k$ is approximately $6\sqrt{\hat{\lambda}_k}$. So we can control $\sqrt{\hat{\lambda}_k}=0.7\sqrt{\lambda_k}$, then $\sqrt{\lambda_k}-\sqrt{\hat{\lambda}_k}=0.3\sqrt{\lambda_k}$, $k=1,2,\ldots,m$.

Based on the above discussion, we can summarize how to determine the value of the hyperparameter $\beta$ when $l=m$. If $\mathop{\max}_{1 \leq i \leq m}(0.3\sqrt{\lambda_i})\leq \mathop{\min}_{1 \leq i \leq m}(\sqrt{\lambda_i})$, we can set $\beta=\mathop{\max}_{1 \leq i \leq m}(0.3\sqrt{\lambda_i})$. Otherwise, we set $\beta=\mathop{\min}_{1 \leq i \leq m}(\sqrt{\lambda_i})$.

In practical applications, we usually encounter this type of dataset whose intrinsic dimension $l$ is smaller than the actual dimension $m$. In this case, there are only $l$ eigenvalues of the covariance matrix of the dataset that are not close to $0$, we denote them as $\lambda^{'}_k$, $k=1,2,\ldots,l$. In the training process, we only need to control such $l$ eigenvalues and make $\sqrt{\lambda^{'}_k}-\sqrt{\hat{\lambda}^{'}_k}=0.3\sqrt{\lambda^{'}_k}$, $k=1,2,\ldots,l$. Besides, since $\lambda^{'}_k>0$, $\beta$ should satisfy $0 \leq \beta \leq \mathop{\min}_{1 \leq i \leq l}(\sqrt{\lambda^{'}_i})$. Actually, if $l<m$, there will be some data points that have a large impact on the reconstruction ability of the autoencoder (i.e., IP), it is very important to not only improve the detection effect of HLP, but also maintain the detection effect of IP. Therefore, we have to make sure that the value of $\beta$ is small enough but not equal to $0$. For convenience, we also set $0<\beta<\mathop{\min}_{1 \leq i \leq m}(\sqrt{\lambda_i})$ in this case. Finally, the determination scheme for the value of $\beta$ is the same as when $l=m$.

\subsection*{Ethics statement}
Every effort was made to minimize the number of animals used, and their suffering, the guidelines are available at the IMPC portal: \href{https://www.mousephenotype.org/about-impc/arrive-guidelines}{https://www.mousephenotype.org/about-impc/arrive-guidelines}.

\section*{Data availability}
The raw data from IMPC was downloaded from \href{https://ftp.ebi.ac.uk/pub/databases/impc/all-data-releases/release-13.0/cores/statistical-raw-data_20201217_163937.tar}{https://ftp.ebi.ac.uk/pub/databases/impc/all-data-releases/release-13.0/cores/statistical-raw-data\_20201217\_163937.tar}. The Dry Bean Dataset is available at \href{https://archive.ics.uci.edu/ml/datasets/Dry+Bean+Dataset}{https://archive.ics.uci.edu/ml/datasets/Dry+Bean+Dataset}. The Breast Cancer Dataset is available at \href{https://www.kaggle.com/datasets/uciml/breast-cancer-wisconsin-data}{https://www.kaggle.com/datasets/uciml/breast-cancer-wisconsin-data}.

\section*{Code availability}
The implementation of the DAGMM algorithm is available at \href{https://github.com/tnakae/DAGMM}{https://github.com/tnakae/DAGMM}, and the R code for Cross-Phenotype Meta-Analysis is available at \href{https://rdrr.io/cran/omics/man/cpma.html}{https://rdrr.io/cran/omics/man/cpma.html}. Besides, the main code used for analysis and the trained models will be made available on request.

%
%
%
%
%
%
%
\bibliography{ODBAE_manuscript}
%


\section*{Acknowledgements}
This work was supported by the National Key Research and Development Program of China (2018YFA0801103), the National Natural Science Foundation of China (12071330) to Prof. Ling Yang, the National Key Research and Development Program of China (2018YFA0801101), Interdisciplinary basic Frontier Innovation Program of Suzhou Medical College of Soochow University (YXY2303022), National Center for International Research (2017B01012), the National Natural Science Foundation of China (32100931), Jiangsu Province Engineering Research Center of Development and Translation of Key Technologies for Chronic Disease Prevention and Control, and the Postgraduate Research \& Practice Innovation Program of Jiangsu Province KYCX23\_3225. The authors would like to thank Professor Moli Huang's lab members (School of Biology and Basic Medical Sciences, Soochow University), Professor Yu Tang (School of Mathematical Sciences, Soochow University) and Professor Huanfei Ma (School of Mathematical Sciences, Soochow University) for their valuable suggestions and kind help.

\section*{Author contributions statement}
Y.S., Y.X. and L.Y. designed the research. Y.S. performed framework implementation and generated results. Y.S., T.Z., Z.L. and K.K. analyzed the data and interpreted the results. L.Y. and Y.X. provided overall supervision and direction of the work. Manuscript writing was done by Y.S. and T.Z.
 
\section*{Competing interests}
The authors declare no competing interests.
\newpage

\setcounter{figure}{0}
\section*{Supplementary figures}
\captionsetup[figure]{labelfont={bf},name={Supplementary Figure}}

\begin{figure}[H]
	\centering %
	\subfigure{
		\label{section1_PCA_train_data}
		\begin{overpic}[width=0.48\textwidth]{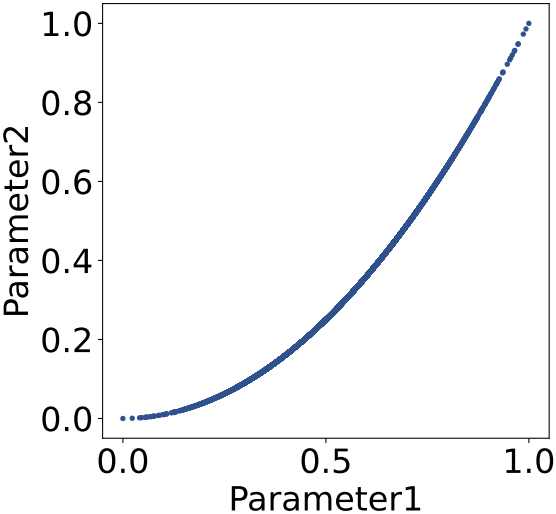}
			\put(1,90){\large\textbf{a}}
	\end{overpic}}
	\subfigure{
		\label{section1_PCA_test_data}
		\begin{overpic}[width=0.48\textwidth]{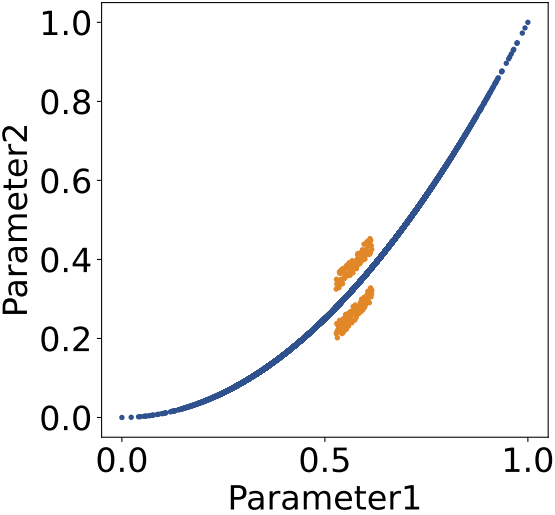}
			\put(1,90){\large\textbf{b}}
	\end{overpic}}

	\subfigure{
		\label{section1_PCA_PCA_200}
		\begin{overpic}[width=0.48\textwidth]{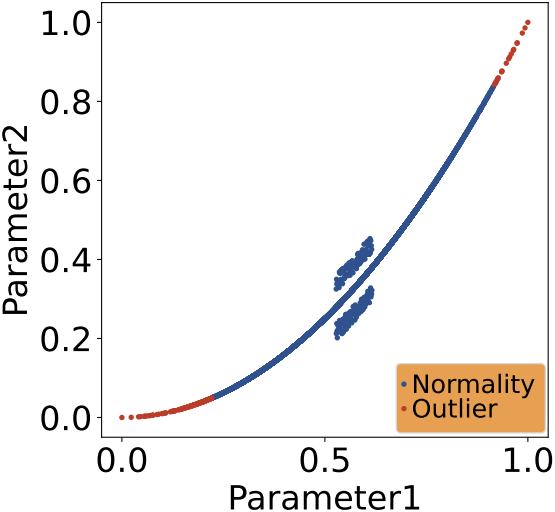}
			\put(1,90){\large\textbf{c}}
	\end{overpic}}
	\subfigure{
		\label{section1_PCA_mse_loss_200}
		\begin{overpic}[width=0.48\textwidth]{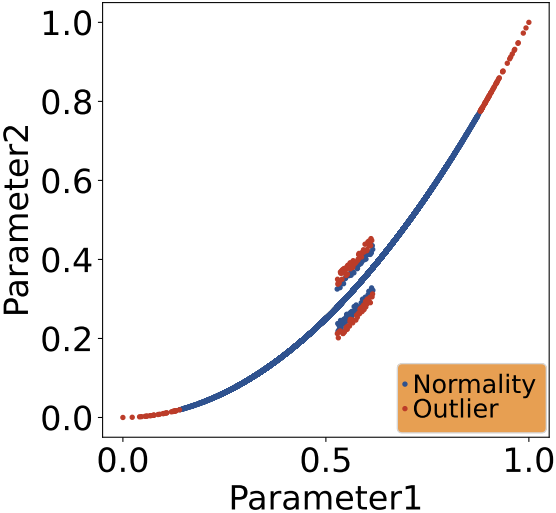}
			\put(1,90){\large\textbf{d}}
	\end{overpic}}
	\caption{{\bf Datasets with nonlinear features and their corresponding outlier detection results by PCA and autoencoder.} {\bf a}, Training dataset, the value of Parameter2 is equal to the square of the corresponding value of Parameter1. {\bf b}, Test dataset,  the blue data points have the same nonlinear relationship between dimensions as the training dataset, while the red data points do not. {\bf c}, Outlier detection result of PCA. {\bf d}, Outlier detection result of autoencoder.}
	\label{ext:fig1}
\end{figure}

\begin{figure}[H]
	\centering
	\subfigure{
		\label{section1_generate_example1_mse_loss1000_tanh_none}
		\begin{overpic}[width=0.48\textwidth]{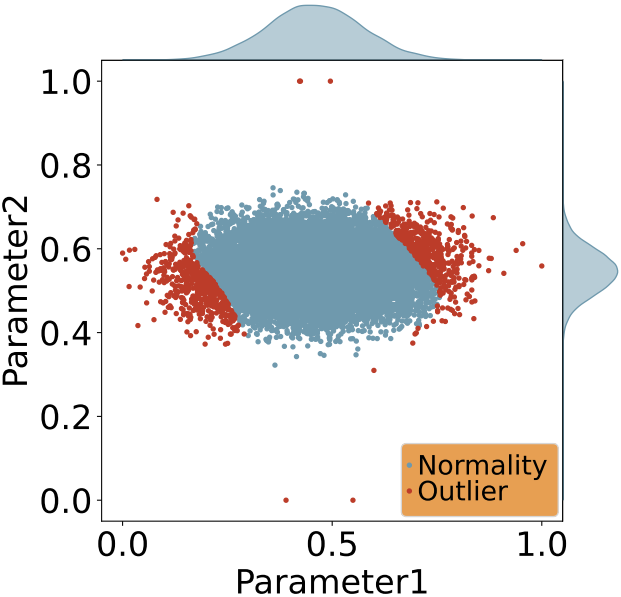}
			\put(1,90){\large\textbf{a}}
	\end{overpic}}
	\subfigure{
		\label{section1_generate_example2_mse_loss1000_tanh_none}
		\begin{overpic}[width=0.48\textwidth]{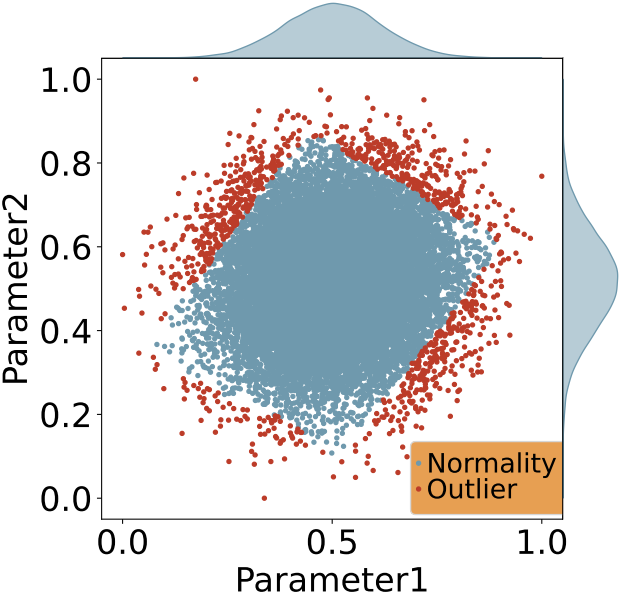}
			\put(1,90){\large\textbf{b}}
	\end{overpic}}
	
	\subfigure{
		\label{section1_generate_example3_mse_loss1000_tanh_none}
		\begin{overpic}[width=0.48\textwidth]{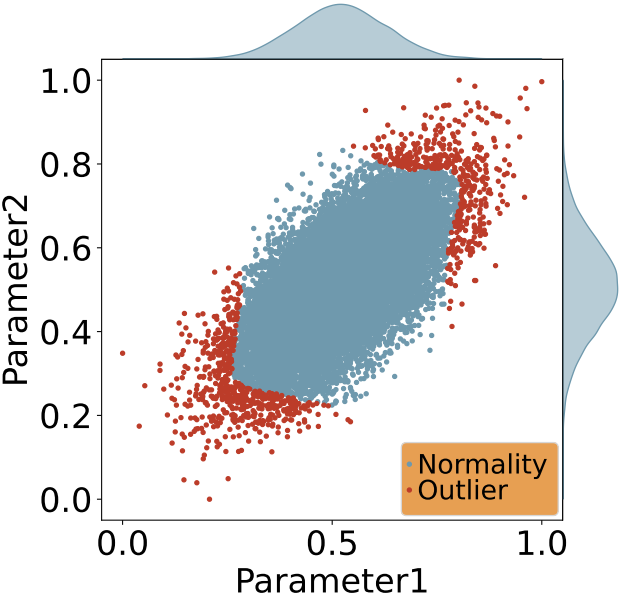}
			\put(1,90){\large\textbf{c}}
	\end{overpic}}
	\subfigure{
		\label{section1_generate_example4_mse_loss1000_tanh_none}
		\begin{overpic}[width=0.48\textwidth]{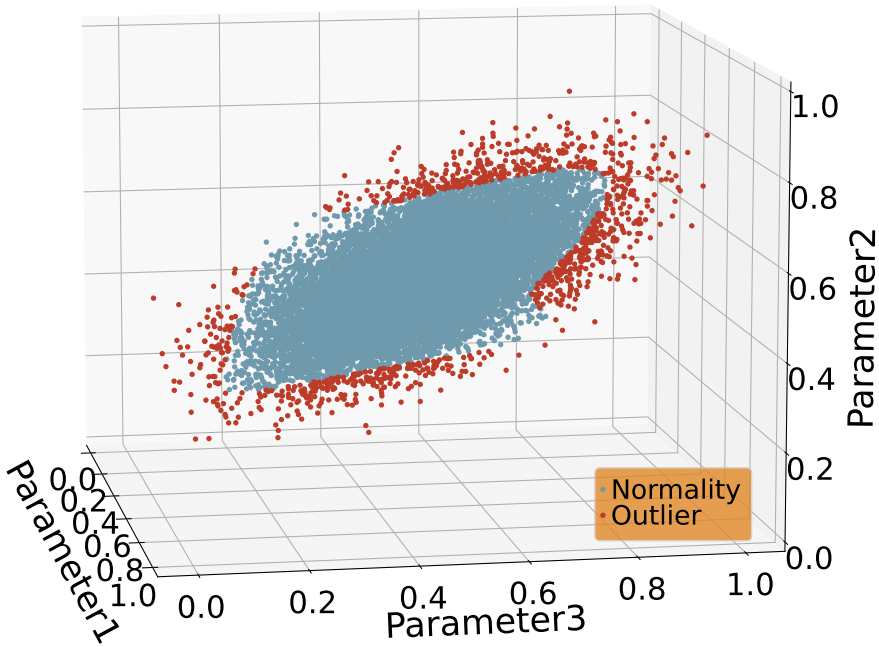}
			\put(1,90){\large\textbf{d}}
	\end{overpic}}
	\caption{Outlier detection results for all datasets in Fig.2c-2e, 2i when the activation functions of the hidden layer in the autoencoder are tanh, and no activation function is applied to the output layer.}
	\label{ext:fig3}
\end{figure}

\begin{figure}[H]
	\centering
	\subfigure{
		\label{section1_generate_example1_mse_loss1000_tanh_sigmoid}
		\begin{overpic}[width=0.48\textwidth]{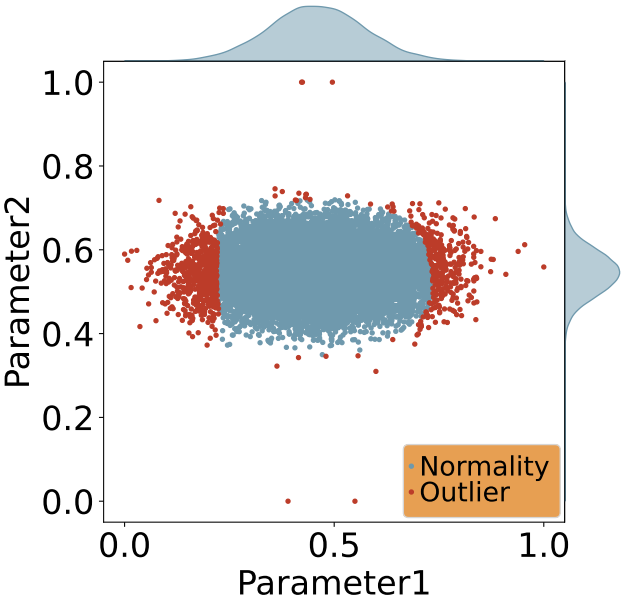}
			\put(1,90){\large\textbf{a}}
	\end{overpic}}
	\subfigure{
		\label{section1_generate_example2_mse_loss1000_tanh_sigmoid}
		\begin{overpic}[width=0.48\textwidth]{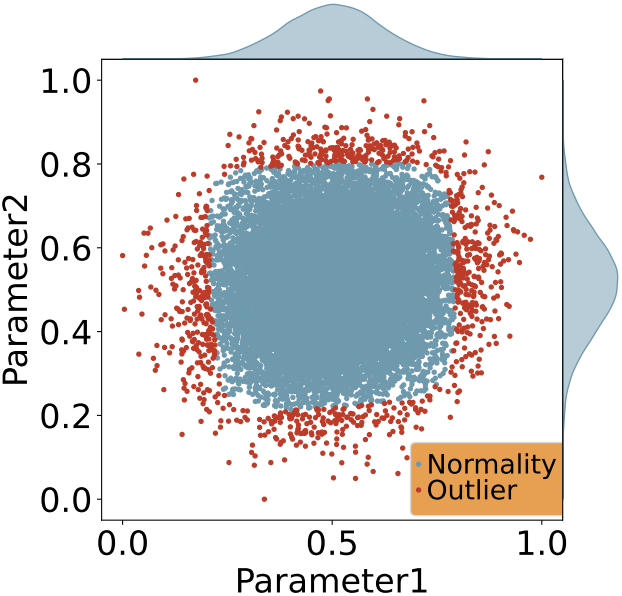}
			\put(1,90){\large\textbf{b}}
	\end{overpic}}
	
	\subfigure{
		\label{section1_generate_example3_mse_loss1000_tanh_sigmoid}
		\begin{overpic}[width=0.48\textwidth]{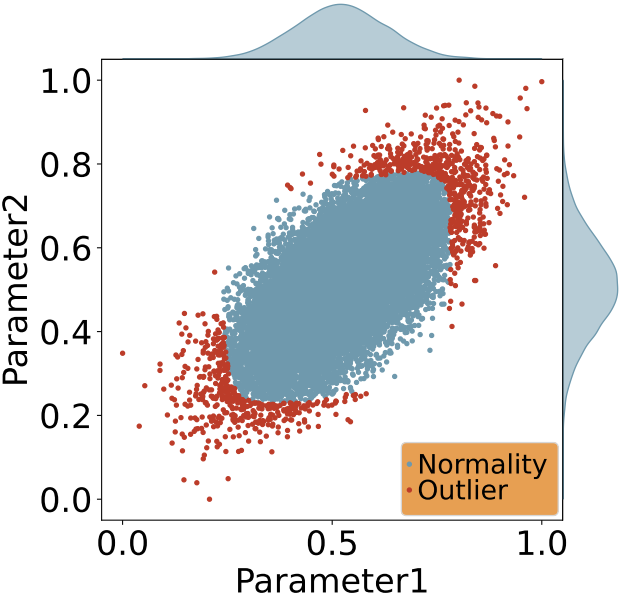}
			\put(1,90){\large\textbf{c}}
	\end{overpic}}
	\subfigure{
		\label{section1_generate_example4_mse_loss1000_tanh_sigmoid}
		\begin{overpic}[width=0.48\textwidth]{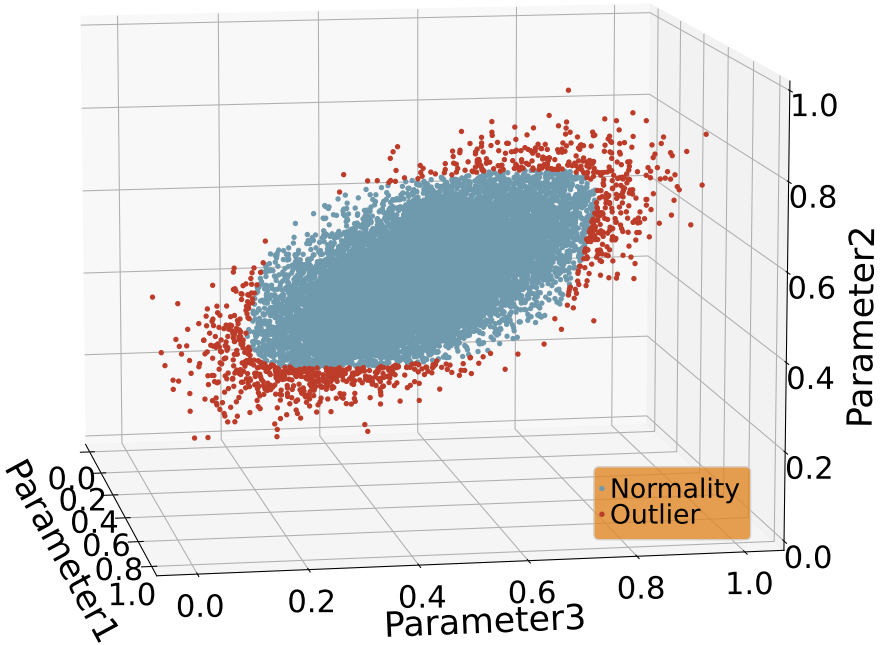}
			\put(1,90){\large\textbf{d}}
	\end{overpic}}
	\caption{Outlier detection results for all datasets in Fig.2c-2e, 2i when the activation functions of the hidden layer and output layer of the autoencoders are tanh and sigmoid, respectively.}
	\label{ext:fig4}
\end{figure}

\begin{figure}[H]
	\centering
	\subfigure{
		\label{section1_generate_example2_mse_input_output}
		\begin{overpic}[width=0.48\textwidth]{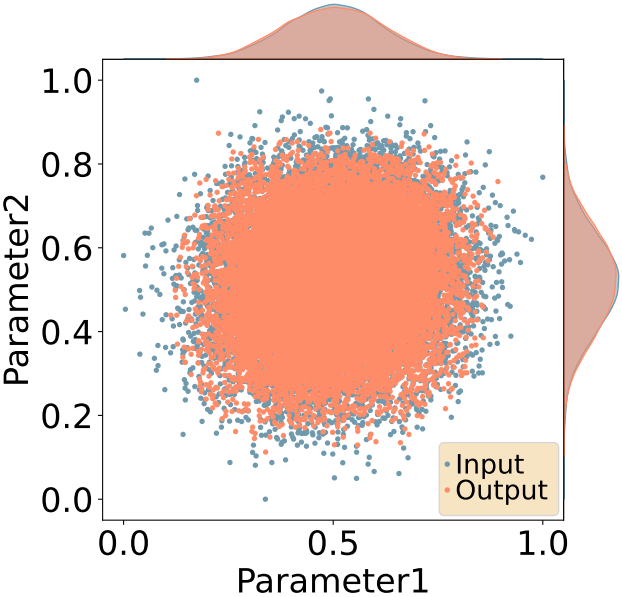}
			\put(1,90){\large\textbf{a}}
	\end{overpic}}
	\subfigure{
		\label{section1_generate_example2_mse_eig_input_output}
		\begin{overpic}[width=0.48\textwidth]{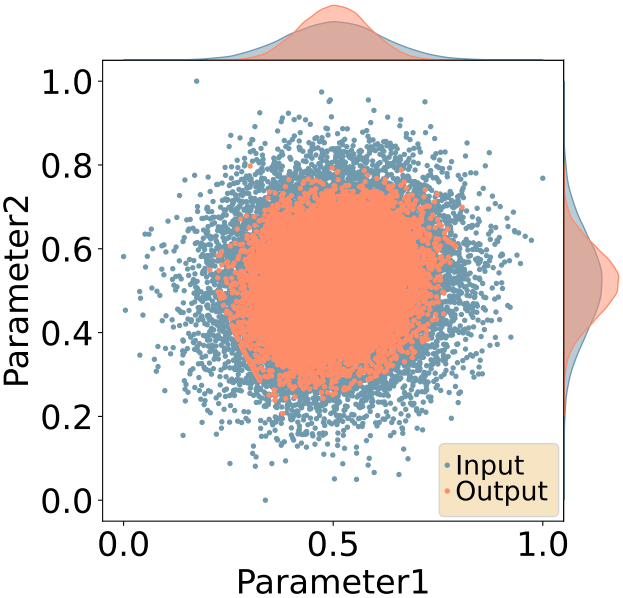}
			\put(1,90){\large\textbf{b}}
	\end{overpic}}
	
	\subfigure{
		\label{section1_generate_example2_reconstruction_error_mse}
		\begin{overpic}[width=0.48\textwidth]{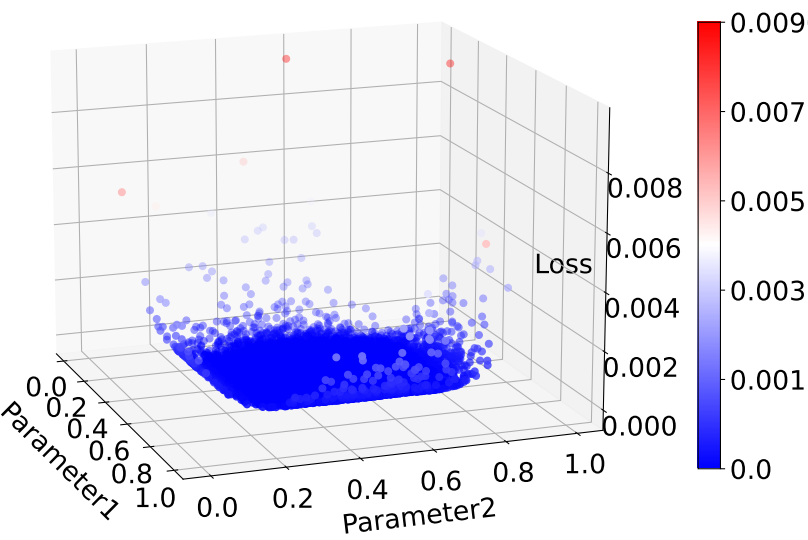}
			\put(1,65){\large\textbf{c}}
	\end{overpic}}
	\subfigure{
		\label{section1_generate_example2_reconstruction_error_SCRAE}
		\begin{overpic}[width=0.48\textwidth]{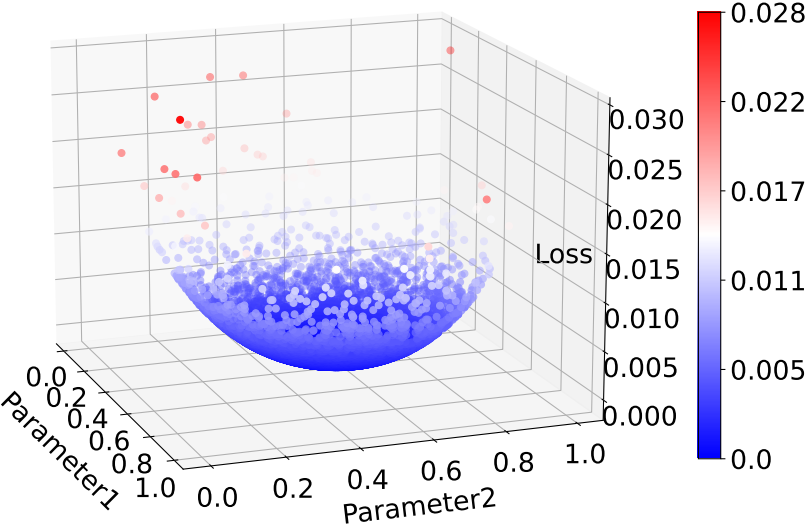}
			\put(1,65){\large\textbf{d}}
	\end{overpic}}
	\caption{{\bf Reconstruction results and reconstruction errors of datasets in Fig.2d and Fig.2g.} {\bf a}, Reconstruction result corresponding to Fig.2d, the distribution of reconstructed dataset is independent across two dimensions. {\bf b}, Reconstruction result corresponding to Fig.2g, the reconstructed dataset follows a balanced joint distribution. {\bf c}, Reconstruction errors corresponding to Fig.2d, HLP in each dimension is independently detected by MSE-trained autoencoder. {\bf d}, Reconstruction errors corresponding to Fig.2g, all HLP reconstructed through ODBAE will produce large reconstruction errors and then be identified.}
	\label{ext:fig10}
\end{figure}

\begin{figure}[H]
	\centering
	\subfigure{
		\label{section4_generate_example4_train_data}
		\begin{overpic}[width=0.46\textwidth]{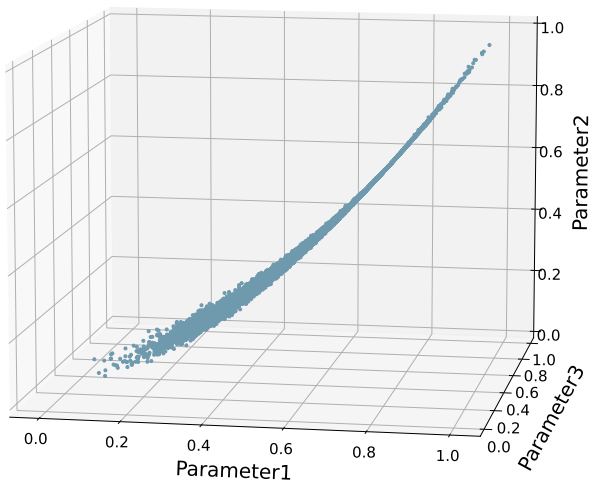}
			\put(2,75){\large\textbf{a}}
	\end{overpic}}
	\hspace{0.1in}
	\subfigure{
		\label{section4_generate_example4_test_data}
		\begin{overpic}[width=0.46\textwidth]{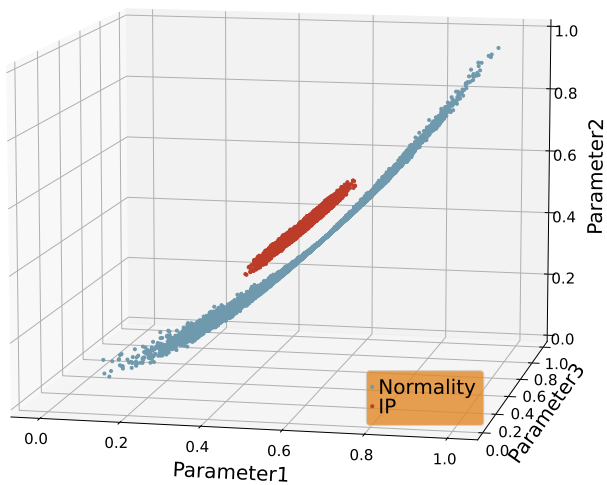}
			\put(-8,75){\large\textbf{b}}
	\end{overpic}}
	
	\subfigure{
		\label{section4_generate_example4_auc_score_combine}
		\begin{overpic}[width=0.48\textwidth]{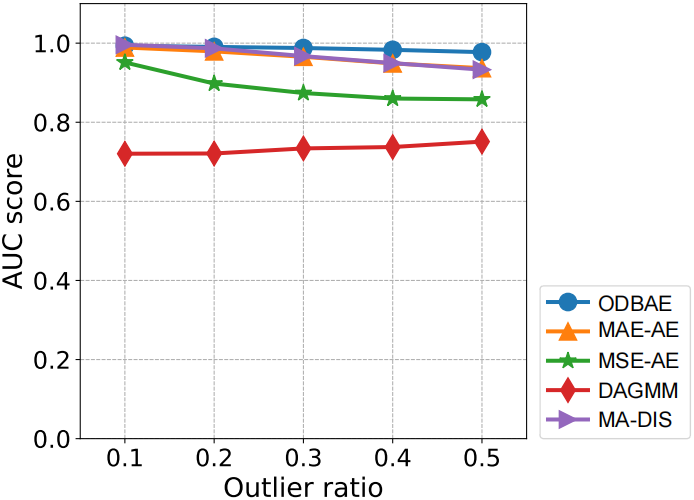}
			\put(2,70){\large\textbf{c}}
	\end{overpic}}
	\subfigure{
		\label{section4_generate_example4_ap_score_combine}
		\begin{overpic}[width=0.48\textwidth]{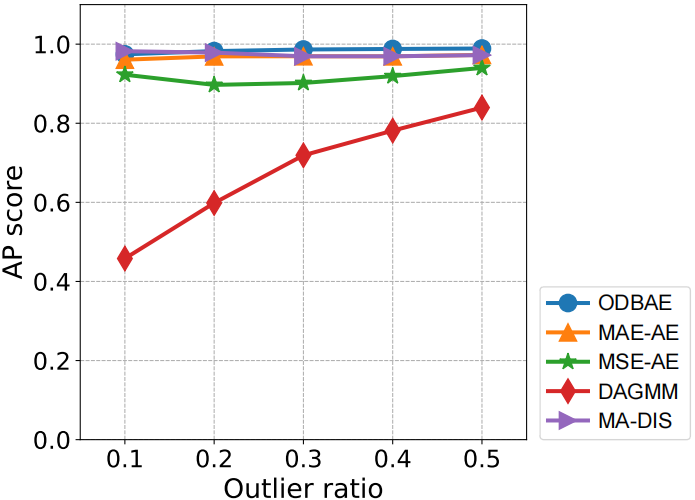}
			\put(2,70){\large\textbf{d}}
	\end{overpic}}		
	\caption{{\bf Comprehensive detection performance of ODBAE for IP and HLP.} {\bf a}, Synthetic training dataset, the value of Parameter2 is equal to the square of the value of Parameter1. {\bf b}, Synthetic test dataset, the correlation between the dimensions of the blue sample points is the same as the training dataset, while the correlation between the dimensions of the red sample points is different from the training dataset. {\bf c},{\bf d}, Comparison results when IP exist in the test dataset and both IP and HLP are considered as positive and we assume the same ratio of IP and HLP.}
	\label{ext:fig5}
\end{figure}

\begin{figure}[H]
	\centering
	\subfigure{
		\label{section4_generate_example5_auc_score_50dim}
		\begin{overpic}[width=0.48\textwidth]{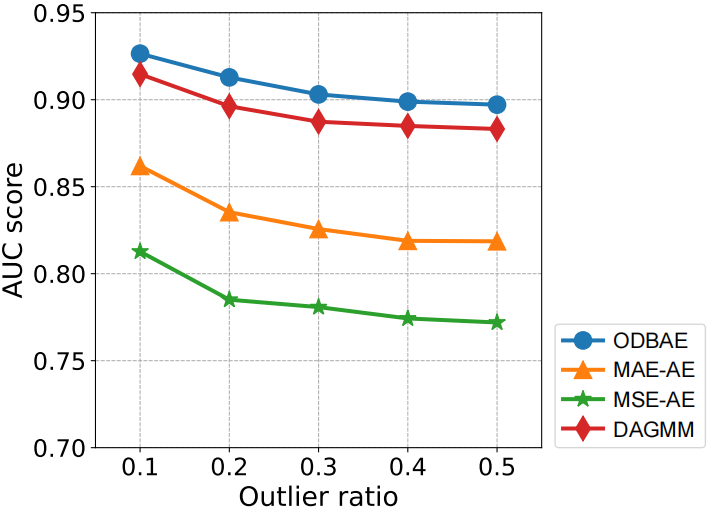}
			\put(2,75){\large\textbf{a}}
	\end{overpic}}
	\subfigure{
		\label{section4_generate_example5_ap_score_50dim}
		\begin{overpic}[width=0.48\textwidth]{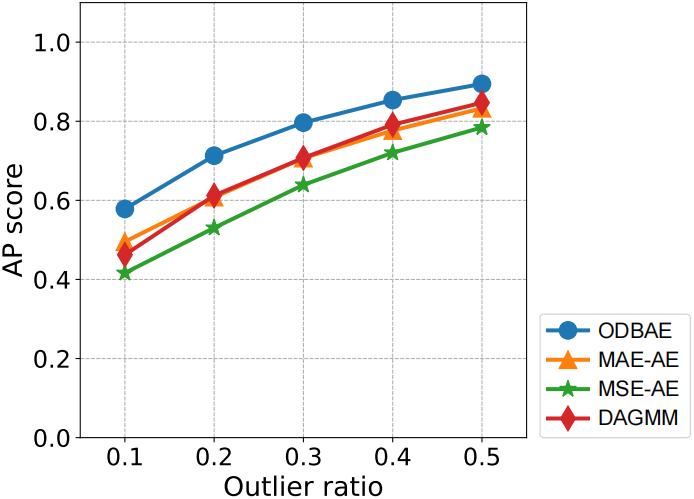}
			\put(2,75){\large\textbf{b}}
	\end{overpic}}

	\subfigure{
		\label{section4_generate_example5_auc_score_100dim}
		\begin{overpic}[width=0.48\textwidth]{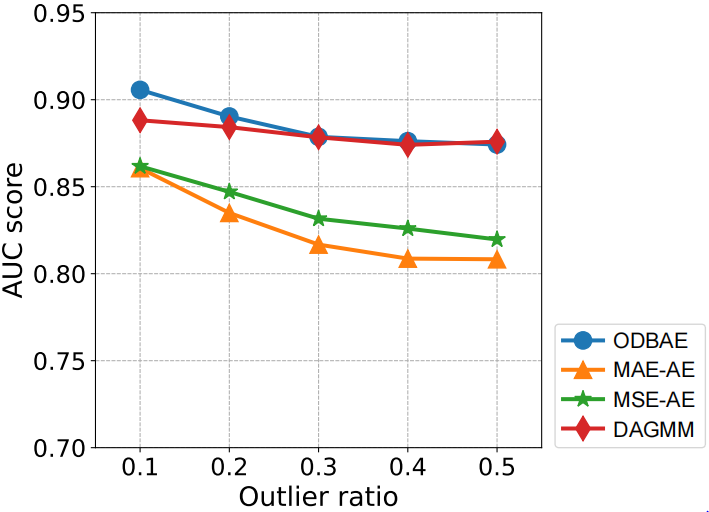}
			\put(2,75){\large\textbf{c}}
	\end{overpic}}
	\subfigure{
		\label{section4_generate_example5_ap_score_100dim}
		\begin{overpic}[width=0.48\textwidth]{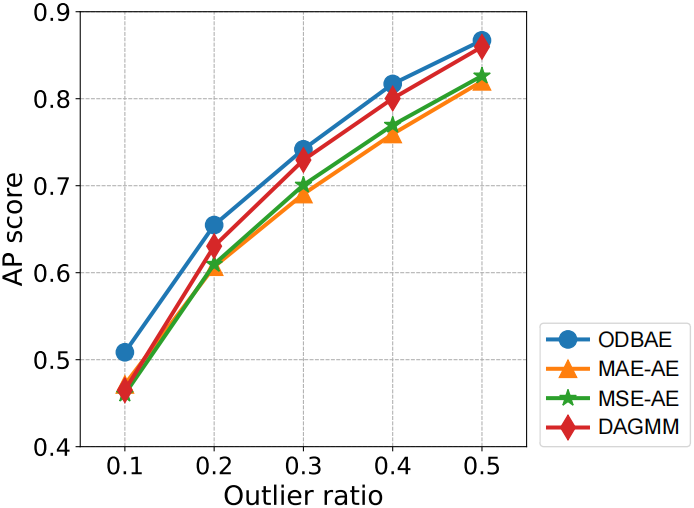}
			\put(2,75){\large\textbf{d}}
	\end{overpic}}
	\caption{{\bf AUC and AP scores for outlier detection result for high-dimensional datasets}. Both datasets synthesized follow a multi-dimensional Gaussian distribution, and their covariance matrices are diagonal. {\bf a},{\bf b}, Outlier detection result for a synthetic $50$-dimensional Gaussian distribution dataset. {\bf c},{\bf d}, Outlier detection result for a synthetic $100$-dimensional Gaussian distribution dataset.}
	\label{ext:fig6}
\end{figure}

\begin{figure}[H]
	\centering
	\subfigure{
		\label{Alb_HDLC_Pcdhb10}
		\begin{overpic}[width=0.48\textwidth]{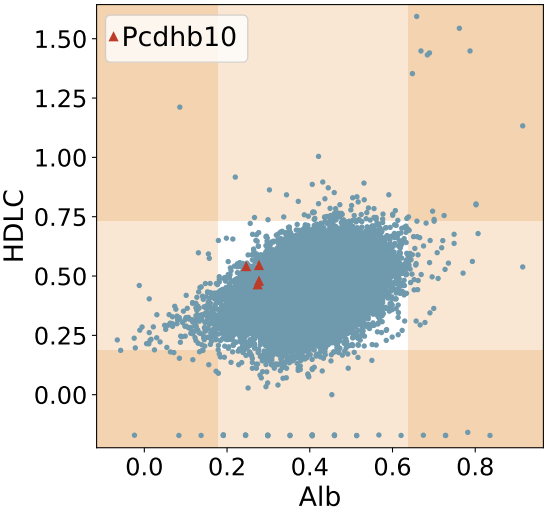}
			\put(1,90){\large\textbf{a}}
	\end{overpic}}
	\subfigure{
		\label{Alb_HDLC_Shisa4}
		\begin{overpic}[width=0.48\textwidth]{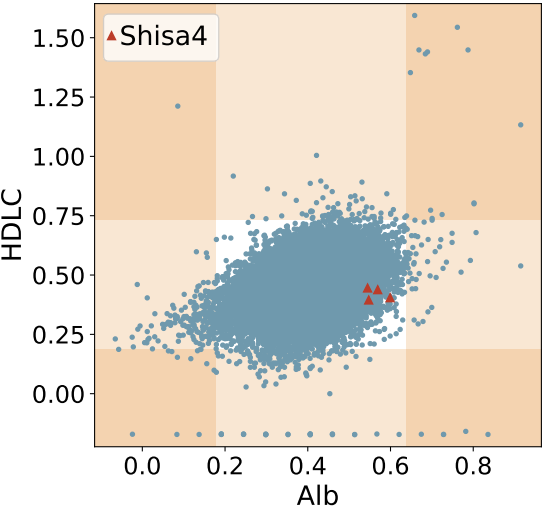}
			\put(1,90){\large\textbf{b}}
	\end{overpic}}
	
	\subfigure{
		\label{TB_TP_Ankrd13c}
		\begin{overpic}[width=0.48\textwidth]{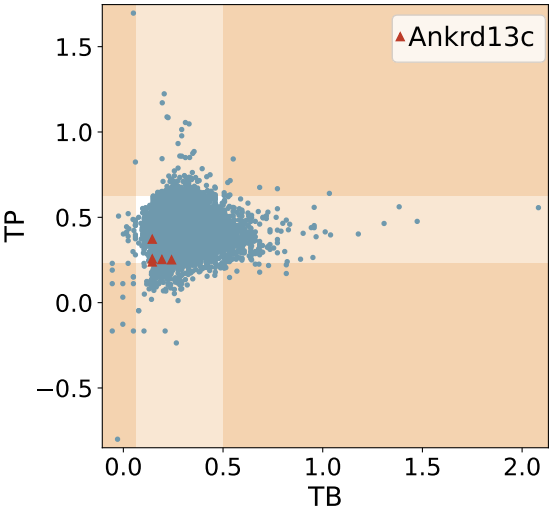}
			\put(1,90){\large\textbf{c}}
	\end{overpic}}
	\subfigure{
		\label{TB_TP_Fndc3a}
		\begin{overpic}[width=0.48\textwidth]{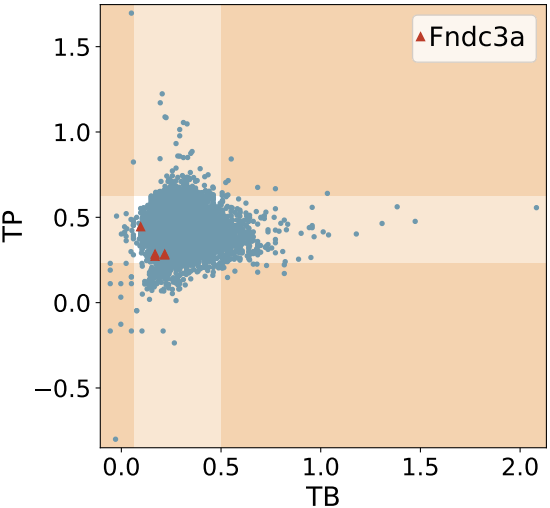}
			\put(1,90){\large\textbf{d}}
	\end{overpic}}
	\caption{{\bf Visualization of outliers corresponding to $4$ novel metabolic genes.} 
		The outliers corresponding to $4$ genes are indeed distributed on the edges of the entire dataset. However, they cannot be detected by the z-score method. The absolute z-score values of at least one parameters of the data points in the shaded part are ranked in the top $1.2$\%.}
	\label{ext:fig7}
\end{figure}

\begin{figure}[H]
	\centering
	\subfigure{
		\label{Alb_Ph_Hoxa10}
		\begin{overpic}[width=0.47\textwidth]{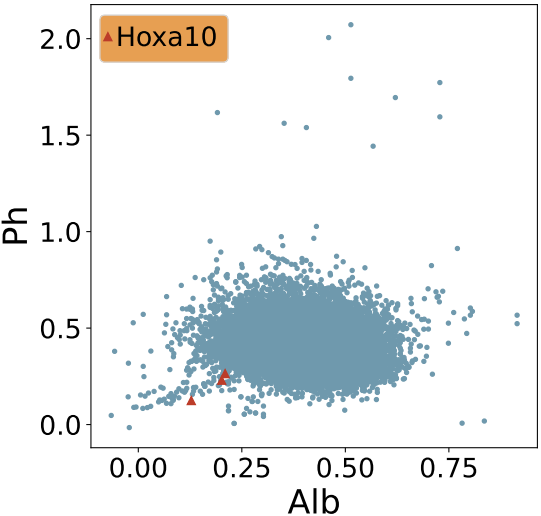}
			\put(1,90){\large\textbf{a}}
	\end{overpic}}
	\subfigure{
		\label{Alb_HDLC_Twf2}
		\begin{overpic}[width=0.5\textwidth]{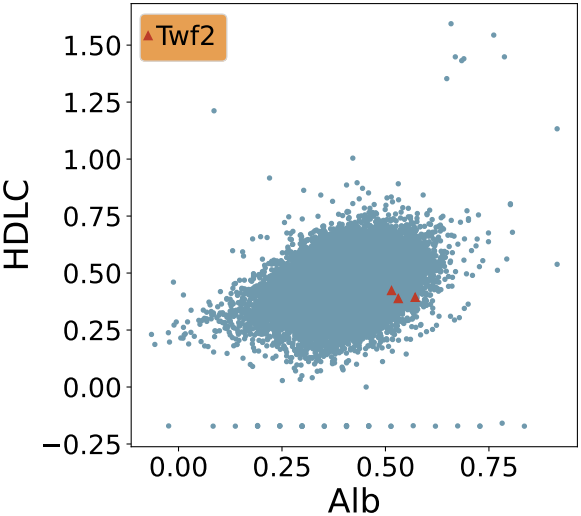}
			\put(1,90){\large\textbf{b}}
	\end{overpic}}
	
	\subfigure{
		\label{ALA_TC_Tmed10}
		\begin{overpic}[width=0.48\textwidth]{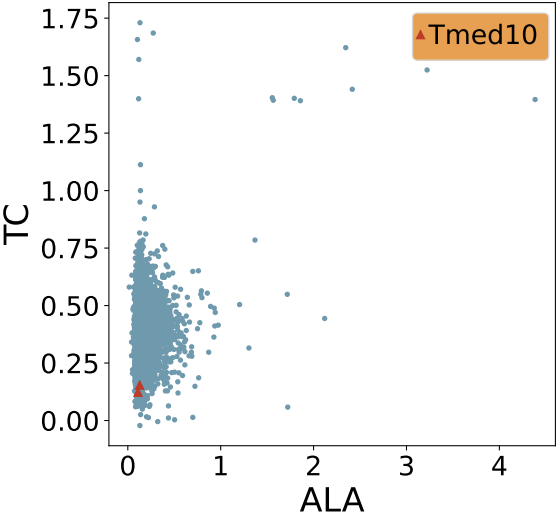}
			\put(1,90){\large\textbf{c}}
	\end{overpic}}
	\hspace{0.03in}
	\subfigure{
		\label{ASA_HDLC_Nbas}
		\begin{overpic}[width=0.47\textwidth]{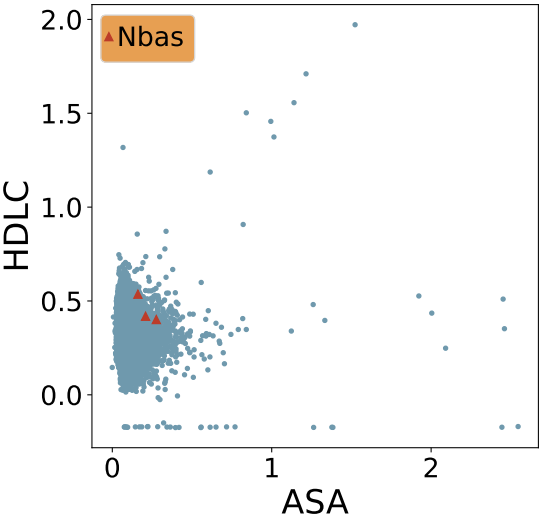}
			\put(1,93){\large\textbf{d}}
	\end{overpic}}
	\caption{{\bf Visualization of outliers corresponding to $4$ genes whose human orthologues with metabolic relevance based on abnormal parameters.} These outliers deviate from the center of all data points, and their corresponding knockout genes are likely to be metabolically related.}
	\label{ext:fig9}
\end{figure}

\begin{figure}[H]
	\centering
	\subfigure{
		\label{ALA_ASA_human_pathway}
		\begin{overpic}[width=0.48\textwidth]{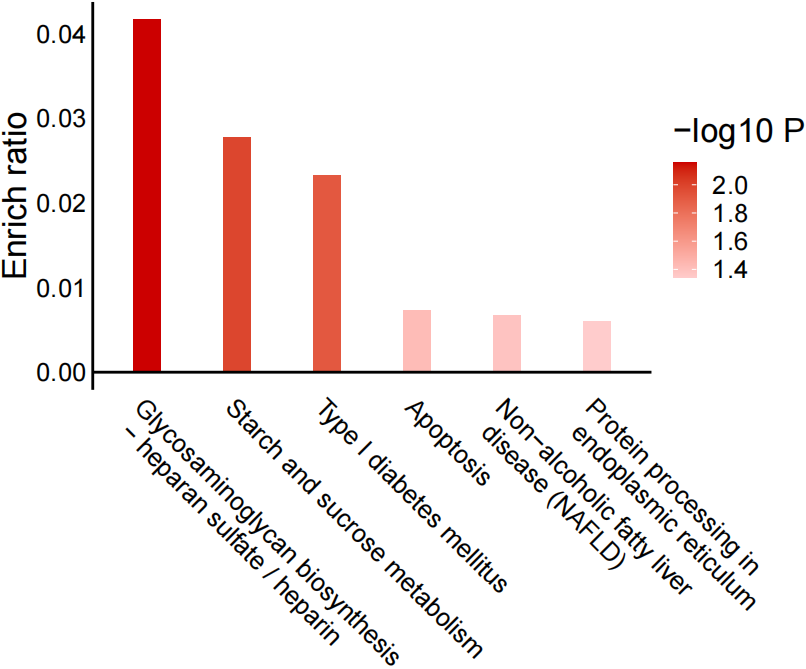}
			\put(-2,85){\large\textbf{a}}
	\end{overpic}}
	\subfigure{
		\label{HDLC_TP_Taf8}
		\begin{overpic}[width=0.48\textwidth]{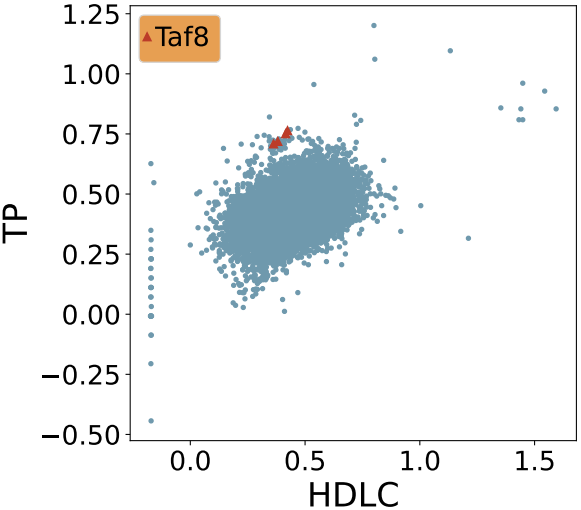}
			\put(-2,85){\large\textbf{b}}
	\end{overpic}}
	
	\subfigure{
		\label{HDLC_TP_fit}
		\begin{overpic}[width=0.48\textwidth]{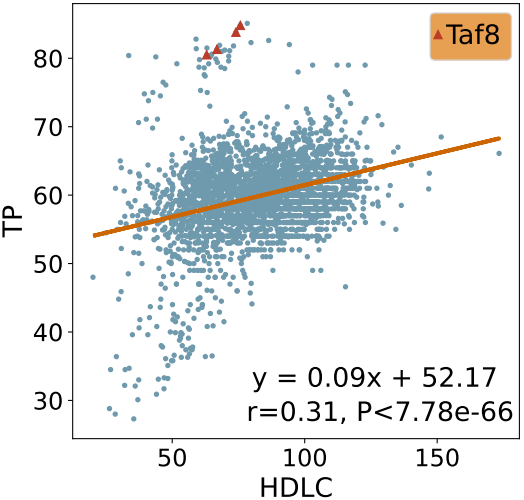}
			\put(-2,85){\large\textbf{c}}
	\end{overpic}}
	\caption{{\bf Study of the internal relationship within abnormal parameter pairs.} {\bf a}, Histogram of KEGG pathway enrichment for genes corresponding to (ALA, ASA) in humans. These genes are enriched in metabolic pathways including Glycosaminoglycan biosynthesis and Starch and sucrose metabolism. {\bf b}, Visualization of outliers corresponding to gene \textit{Taf8} based on abnormal parameters in the male dataset. {\bf c}, Linear regression of parameters HDLC and TP based on the data from the phenotyping center where the gene \textit{Taf8} is located in the male dataset.}
	\label{ext:pathway}
\end{figure}
%
%
%
%

\end{document}